\PassOptionsToPackage{table}{xcolor}
\documentclass[10pt,twocolumn,letterpaper]{article}

\usepackage{cvpr}

\usepackage{array}
\usepackage{multirow}
\usepackage{rotating}
\usepackage{colortbl}
\usepackage{tabularx}
\usepackage{wrapfig}
\usepackage{float}
\usepackage{placeins}
\usepackage{needspace}
\usepackage{cuted}
\usepackage{fix-cm}
\usepackage{etoolbox}

\makeatletter
\patchcmd{\@maketitle}
  {\vskip}
  {\vspace*{-18pt}\vskip}
  {}
  {\PackageWarning{Pros4D}{Title spacing patch failed}}
\makeatother

\def\confName{CVPR}
\def\confYear{2027}

\providecommand{\papertext}[2]{#1}

\ifdefined\ProsChinese
  \usepackage{xeCJK}
  \setCJKsansfont{SimHei}
  \setCJKmonofont{FangSong}

  \renewcommand{\papertext}[2]{#2}
  
  \renewenvironment{abstract}{%
    \vskip.075in
    \centerline{\large 摘要}
    \vspace{0.5ex}
    \begin{quote}%
  }{%
    \par\end{quote}\vskip 1ex%
  }

\fi

\definecolor{cvprblue}{rgb}{0.21,0.49,0.74}
\usepackage[
  pagebackref,
  breaklinks,
  colorlinks,
  allcolors=cvprblue
]{hyperref}

\newsavebox{\maintablebox}

\definecolor{temporalgray}{HTML}{F1F1F1}
\definecolor{numcolor}{HTML}{E2ECEF}
\definecolor{qacolor}{HTML}{E1E8F3}
\definecolor{featurecolor}{HTML}{E5E1F5}
\definecolor{numtext}{HTML}{00536B}
\definecolor{qatext}{HTML}{324779}
\definecolor{featuretext}{HTML}{5A3477}
\definecolor{gaincolor}{HTML}{008000}
\definecolor{losscolor}{HTML}{C00000}

\newcolumntype{Y}{>{\centering\arraybackslash}X}

\newcommand{\rotmetric}[1]{%
  \makebox[\hsize][c]{%
    \rotatebox[origin=c]{70}{\textbf{#1}}%
  }%
}

\newcommand{\groupheader}[1]{%
  \rowcolor{blue!8}%
  \multicolumn{8}{@{}l@{}}{\textit{\textbf{#1}}} \\
}

\newlength{\ProsOriginalStripSep}

\title{%
  \papertext
  {Pros4D: Modeling Evolving 4D Worlds in Latent Space for \\
   Dynamic Spatial Reasoning}
  {Pros4D：在潜空间中建模演化的四维世界以实现动态空间推理}%
}

\author{%
  {\normalsize
    Rongxin Gao$^{1,*}$,
    Dongxuan Liu$^{1}$,
    Chu Li$^{2}$,
    Chengyang Ji$^{2}$,
    Jialing Huang$^{1}$,
    Runyu Chen$^{1}$,
    Kairun Wen$^{3}$,
     Chenxin Li$^{3}$
  }\\
  {\normalsize
    Jingrui Ye$^{4}$,
    Haosong Zhen$^{5}$,
    Xinghao Ding$^{1,\dagger}$,
    Yue Huang$^{1}$,
    Yuzhi Huang$^{4,\dagger}$,
    Xiaotong Tu$^{1}$,
    Jingyan Jiang$^{4}$%
  }\\[0pt]
  {\small
    $^{1}$XMU \qquad
    $^{2}$CUMT \qquad
    $^{3}$CUHK \qquad
    $^{4}$THU \qquad
    $^{5}$HIT%
  }\\[0pt]
  {\small
    $^{*}$Project Lead \qquad
    $^{\dagger}$Corresponding Authors%
  }\\[0pt]
  {\small
    Project Page:
    \href{https://pros4d.github.io/}{https://pros4d.github.io/}%
  }%
}

\begin{document}

\maketitle

\vspace{-46pt}
\setlength{\ProsOriginalStripSep}{\stripsep}
\setlength{\stripsep}{0pt}

\begin{strip}
  \centering
  \includegraphics[width=\textwidth]{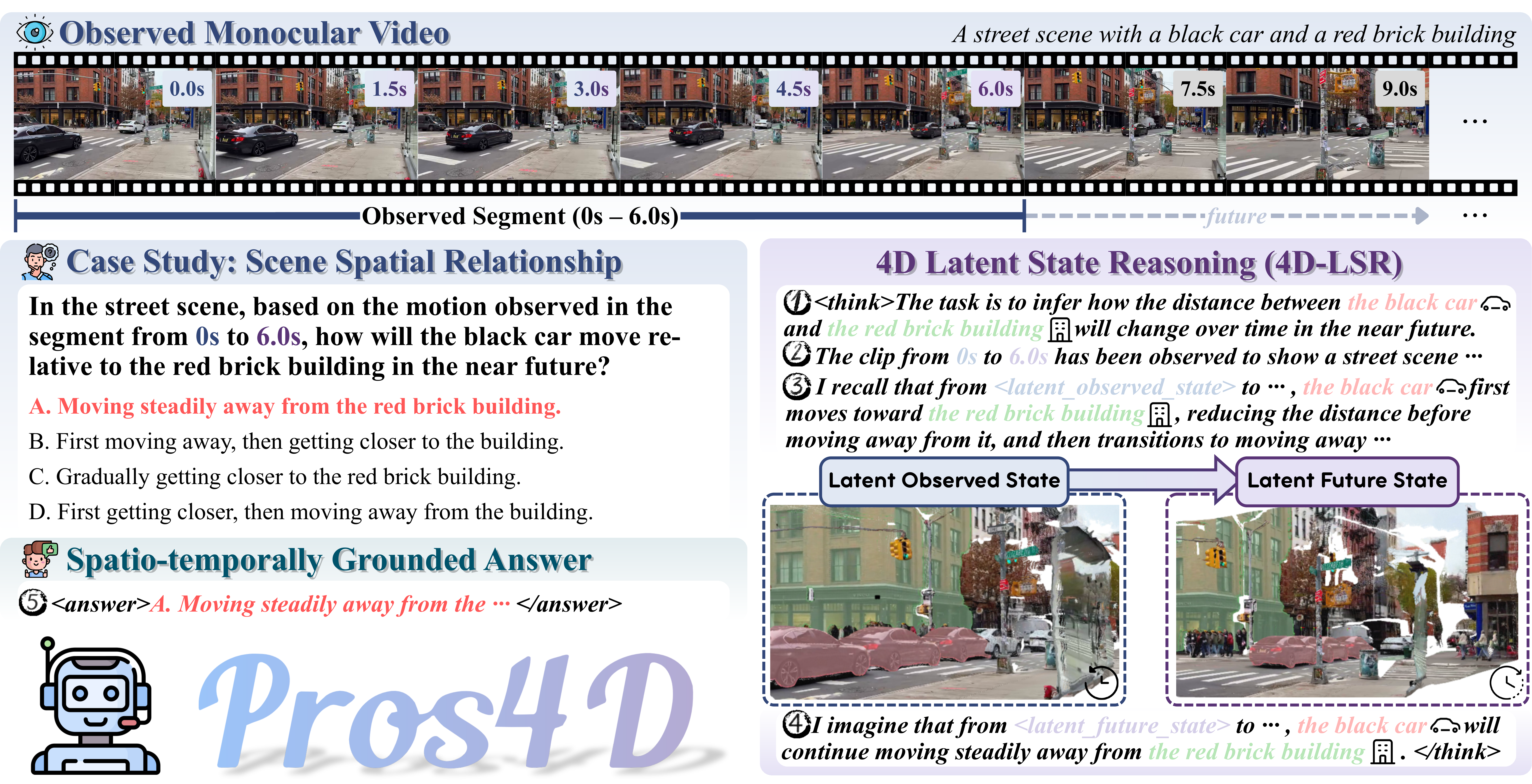}

  \captionsetup{hypcap=false}
  \captionof{figure}{%
    \papertext
    {\textbf{Pros4D} supports observed-dynamics understanding with latent
     observed states and future-dynamics prediction through autoregressive
     rollout of latent future states. Both capabilities reason from the
     same observed video prefix within one VLM at inference.}
    {\textbf{Pros4D} 利用观测潜在状态支持已观测动态理解，并通过自回归推演未来潜在状态实现未来动态预测。两项能力在同一 VLM 内基于共享的观测视频前缀进行推理。}%
  }
  \label{fig:pros4d_teaser}
  \captionsetup{hypcap=true}

  \vspace{5pt}
\end{strip}

\setlength{\stripsep}{\ProsOriginalStripSep}

\begin{abstract}
\papertext{Dynamic spatial reasoning from monocular video enables vision-language models (VLMs) to understand how 3D environments evolve over time, supporting physical world understanding and embodied intelligence. Existing efforts primarily address observed-dynamics understanding, whereas agents in dynamic environments also require future-dynamics prediction beyond the observation boundary. Moreover, external geometric processing and textual descriptions can increase inference complexity or lose information when representing continuous dynamics. These limitations motivate a shared internal state that preserves observed dynamics and remains predictive beyond the input video. Accordingly, we present \textbf{Pros4D}, a framework for modeling evolving 4D worlds in latent space that unifies retrospective understanding of observed-dynamics with prospective reasoning for future-dynamics prediction. Specifically, \textbf{Pros4D-Engine} derives geometry-grounded supervision and constructs paired examples for both tasks, while \textbf{Pros4D-Bench} evaluates them across eight spatial task categories spanning dynamic cameras, objects, and scenes. At the model level, \textbf{Pros4D} compresses observed dynamics into latent observed states and autoregressively rolls them forward into latent future states. \textbf{4D Latent Alignment (4DLA)} grounds these states through complementary visual-semantic and geometric supervision, while joint language supervision connects them to task reasoning. Experiments across benchmarks show consistent improvements over strong baselines in both capabilities.}{单目视频中的动态空间推理使视觉语言模型（VLM）能够理解三维环境如何随时间演化，从而支持物理世界理解与具身智能。现有工作主要关注已观测动态理解，而动态环境中的智能体还需要预测观测边界之外的未来动态。此外，外部几何处理和文本描述在表达连续动态时可能增加推理复杂度或损失信息。这些局限促使我们构建一种共享内部状态，既保留已观测动态，又能对输入视频之外的后续变化进行预测。为此，我们提出 \textbf{Pros4D}，在潜空间中建模演化的四维世界，将对已观测四维动态的回顾性理解与面向未来动态预测的前瞻性推理统一起来。具体而言，\textbf{Pros4D-Engine} 提取基于几何的监督，为两项任务构造成对样本；\textbf{Pros4D-Bench} 则通过覆盖动态相机、物体和场景的八类空间任务进行评测。在模型层面，Pros4D 将已观测动态压缩为观测潜在状态，并以自回归方式向前推演为未来潜在状态。\textbf{四维潜在对齐（4DLA）} 通过互补的视觉语义与几何监督约束这些状态，联合语言监督则将其与任务推理关联。多个基准上的实验表明，Pros4D 在两项能力上均较强基线取得一致提升。}
\end{abstract}

\vspace{-2em}

\section{\papertext{Introduction}{引言}}
\label{secintro}
\papertext{Dynamic spatial reasoning aims to understand how cameras, objects, and their spatial relations evolve over time in a 3D environment \citep{dynbench,dsrsuite,stthink}. From monocular video, this requires an agent to disentangle observer and object motion, preserve spatial relations across time, and reason about how these dynamics may continue beyond the observed segment \citep{future3d,objectforesight}. Such capability is essential for connecting visual perception with embodied decision making, where an agent must not only understand what has happened but also predict what may happen next \citep{world,gr1,seer,vjepa2}. Although vision-language models (VLMs) have made substantial progress in spatial understanding \citep{spatial-mllm,mllm4d,4drgpt,spatialrgpt,spatialreasoner}, extending this ability from observed dynamics to subsequent 4D evolution remains an open challenge for video-based agents.}{动态空间推理旨在理解三维环境中相机、物体及其空间关系如何随时间演化 \citep{dynbench,dsrsuite,stthink}。对于单目视频，智能体需要解耦观察者与物体运动，跨时间保留空间关系，并推理这些动态如何延续到已观测片段之外 \citep{future3d,objectforesight}。这一能力对于连接视觉感知与具身决策至关重要，因为智能体不仅需要理解已发生的变化，还需要预测后续可能发生的情况\citep{world,gr1,seer,vjepa2}。尽管视觉语言模型（VLM）在空间理解方面已取得显著进展 \citep{spatial-mllm,mllm4d,4drgpt,spatialrgpt,spatialreasoner}，将这一能力从已观测动态扩展到后续四维演化对基于视频的智能体而言仍是开放挑战。}

\papertext{Existing 4D reasoning methods primarily focus on observed-dynamics understanding. Explicit approaches augment VLMs with geometric modules or tool-mediated computation \citep{geometrically,spatialclaw,sagent,dyntrace}, or use textual reasoning through prompting \citep{trace} and chain-of-thought fine-tuning \citep{mllm4d,agentdistill,spatialladder,prosr}. Geometric augmentation adds inference-time components and perception dependencies, while verbalizing dynamics can lose fine-grained motion and geometric information. More importantly, supervision centered on observed evidence does not require a state that continues evolving once observation ends. Dynamic spatial reasoning therefore calls for an internal representation that preserves observed dynamics while remaining predictive beyond the input video in dynamic environments.}{现有四维推理方法主要关注已观测动态理解。显式方法通过几何模块或工具计算增强 VLM \citep{geometrically,spatialclaw,sagent,dyntrace}，或通过提示 \citep{trace} 和思维链微调 \citep{mllm4d,agentdistill,spatialladder,prosr} 引导文本推理。几何增强引入额外推理组件及对感知模块的依赖，而将动态语言化可能损失细粒度运动与几何信息。更重要的是，以观测证据为中心的监督并不要求状态在观测结束后继续演化。因此，动态空间推理需要一种内部表征，既保留已观测动态，又能预测动态环境中输入视频之外的后续变化。}

\papertext{An implicit latent state provides a natural basis for such a representation. People integrate successive observations into an internal model to interpret ongoing motion and imagine what may happen next. Latent feature prediction learns visual and motion representations \citep{vjepa} and supports forecasting for control \citep{videoworld,vjepa2,dinowm}. Prior work explores latent visual computation for fine-grained perception \citep{LVR,visreflect} and spatial memory for retaining observed scene structure \citep{vision,latentwm}. Video world models also combine latent dynamics forecasting with semantic guidance from VLMs \citep{thinkjepa}. These advances motivate shared states that preserve observed 4D dynamics and support geometrically grounded evolution beyond the observation boundary in dynamic scenes.}{隐式潜在状态为这样的表征提供了自然基础。人类将连续观测整合为内部模型，用于理解正在发生的运动并想象后续变化。潜在特征预测能够学习视觉与运动表征 \citep{vjepa}，并为控制任务中的预测提供支持 \citep{videoworld,vjepa2,dinowm}。已有工作探索了面向细粒度感知的潜在视觉计算 \citep{LVR,visreflect}，以及保留已观测场景结构的空间记忆 \citep{vision,latentwm}。视频世界模型还将潜在动态预测与 VLM 提供的语义引导结合 \citep{thinkjepa}。这些进展促使我们构建共享状态，既保留已观测四维动态，又以几何为依据支持其在动态场景中超出观测边界的演化。}

\papertext{Following this principle, we present \textbf{Pros4D}, a framework for modeling evolving 4D worlds in latent space that unifies retrospective understanding of observed-dynamics with prospective reasoning for future-dynamics prediction. To support both capabilities within this shared modeling framework, we develop \textbf{Pros4D-Engine}, an automated data engine that derives motion and spatial supervision from estimated 3D geometry and constructs paired examples for the two tasks. We further introduce \textbf{Pros4D-Bench}, which evaluates these tasks across eight spatial categories spanning dynamic cameras, objects, and scenes.}{遵循这一原则，我们提出 \textbf{Pros4D}，在潜空间中建模演化的四维世界，将对已观测四维动态的回顾性理解与面向未来动态预测的前瞻性推理统一起来。为在这一统一建模框架下同时支持这两项能力，我们构建自动化数据引擎 \textbf{Pros4D-Engine}，从估计的三维几何中提取运动与空间关系监督，并为两项任务构建成对样本。在此基础上，我们进一步提出 \textbf{Pros4D-Bench}，通过涵盖动态相机、物体和场景的八类空间任务，对这两项能力进行系统评测。}

\papertext{Within this framework, Pros4D performs \textbf{4D Latent State Reasoning (4D-LSR)} by compressing observed dynamics into continuous latent observed states and autoregressively rolling them forward into latent future states for future-dynamics prediction. \textbf{4D Latent Alignment (4DLA)} grounds both types of state through complementary visual-semantic and geometric supervision from observed and held-out frames, while joint language supervision connects these states to task reasoning. At inference, 4D-LSR operates entirely within the VLM using only the observed video prefix, without external geometric tool calls during question answering. Extensive experiments across multiple dynamic spatial reasoning benchmarks demonstrate consistent improvements over strong baselines in both retrospective understanding and prospective reasoning across 3B and 8B model scales.}{在这一框架中，Pros4D 通过\textbf{四维潜在状态推理（4D-LSR）}将已观测动态压缩为连续的观测潜在状态，再以自回归方式向前推演为未来潜在状态，以实现未来动态预测。\textbf{四维潜在对齐（4DLA）}利用来自观测帧与保留帧的互补视觉语义和几何监督约束两类状态，联合语言监督则将这些状态与任务推理关联。推理时，4D-LSR 仅使用已观测视频前缀，完全在 VLM 内部运行，问答过程中无需调用外部几何工具。多个动态空间推理基准上的广泛实验表明，在 3B 与 8B 两种模型规模下，Pros4D 在回顾性理解与前瞻性推理方面均较强基线取得一致提升。}

\papertext{Our contributions are summarized as follows.}{我们的贡献概括如下。}
\begin{itemize}
    \item \papertext{We propose \textbf{Pros4D}, a unified latent reasoning framework that supports retrospective understanding and prospective reasoning through a shared evolving 4D state.}{我们提出 \textbf{Pros4D}，通过共享的演化四维状态支持回顾性理解与前瞻性推理的潜空间推理框架。}
    \item \papertext{We develop \textbf{Pros4D-Engine}, an automated data engine for paired supervision of observed-dynamics understanding and future-dynamics prediction, together with \textbf{Pros4D-Bench} for evaluating both tasks across eight spatial categories under a shared observation boundary.}{我们构建自动化数据引擎 \textbf{Pros4D-Engine}，为已观测动态理解与未来动态预测提供成对监督，并提出 \textbf{Pros4D-Bench}，在共享观测边界下通过八类空间任务评测两项任务。}
    \item \papertext{We introduce \textbf{4D Latent State Reasoning (4D-LSR)}, which reasons through latent observed and future states, and \textbf{4D Latent Alignment (4DLA)}, which grounds these evolving states with complementary visual-semantic, geometric, and language supervision.}{我们提出\textbf{四维潜在状态推理（4D-LSR）}，通过观测与未来潜在状态进行推理，并提出\textbf{四维潜在对齐（4DLA）}，以互补的视觉语义、几何与语言监督约束这些演化状态。}
\end{itemize}

\section{\papertext{Related Work}{相关工作}}
\label{related_work}

\begin{figure*}[t]
    \centering
    \includegraphics[width=0.96\textwidth]{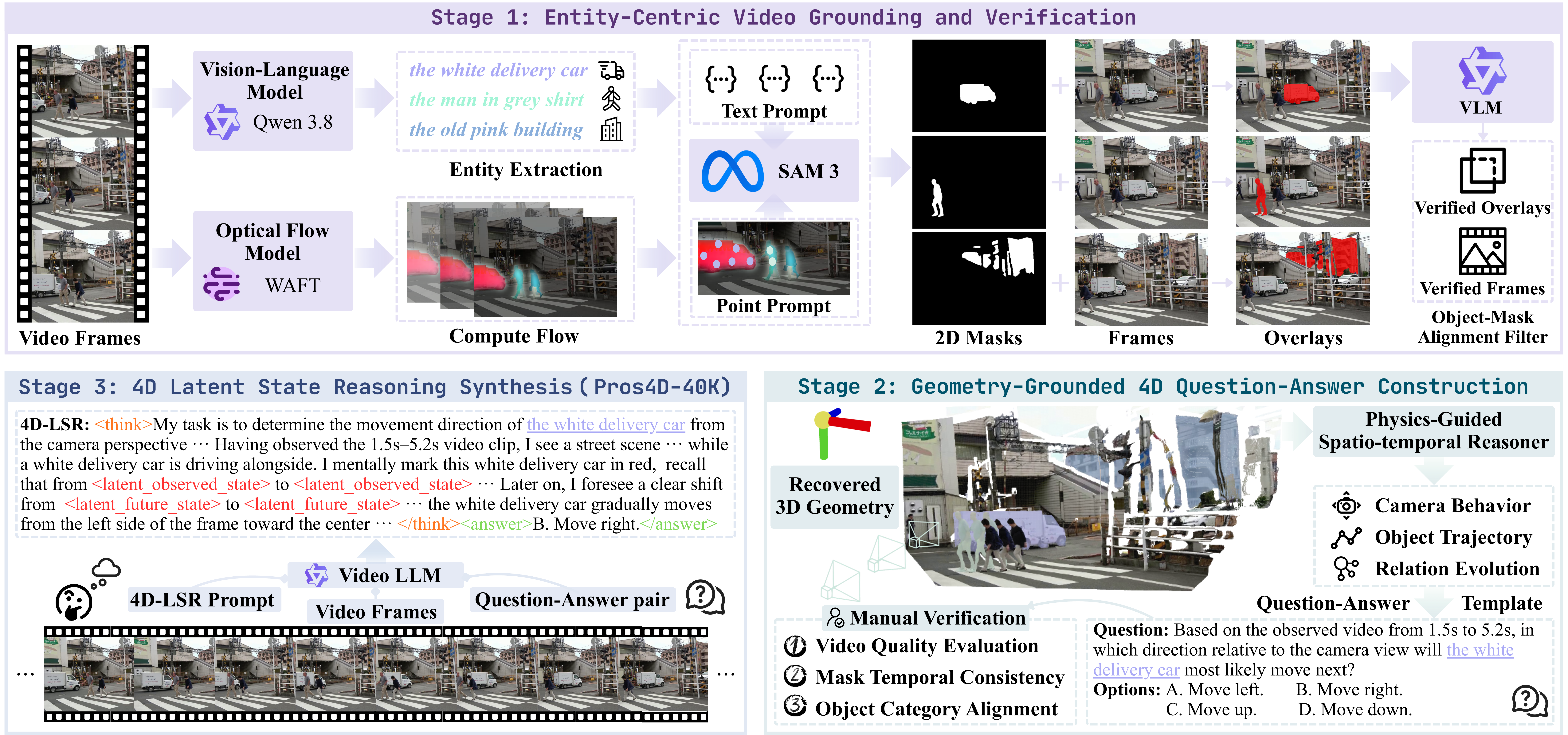}\par
    \vspace{-0.8em}
\caption{\papertext{\textbf{Overview of Pros4D-Engine}. Given raw videos, Stage 1 grounds and verifies entities to obtain reliable object-mask alignments. Stage 2 recovers 3D geometry and constructs question-answer pairs for camera motion, object trajectories, and relation evolution. Stage 3 synthesizes 4D latent-state reasoning traces, pairing textual reasoning with observed- and future-state supervision.}{Pros4D-Engine 概述。给定原始视频，第一阶段对实体进行定位与验证，以获得可靠的对象掩码对齐结果。第二阶段恢复三维几何结构，并构建关于相机运动、物体轨迹及关系演变的问答对。第三阶段合成四维潜在状态推理轨迹，将文本推理与观测状态和未来状态的监督信号相结合。}}
    \label{fig:data_engine}
    \vspace{-1.5em}
\end{figure*}

\subsection{\papertext{Latent Space Reasoning}{潜空间推理}}
\papertext{Latent space reasoning studies how intermediate computation can proceed without verbalizing every step. Textual chains make reasoning explicit but bind it to discrete language \citep{chain}. Work on latent reasoning relaxes this constraint through pause tokens, progressive internalization of reasoning steps, and continuous hidden states \citep{think,explicit,training,codi,softthinking}. These approaches separate computation from its verbal expression. For multimodal problems, intermediate representations must also retain task-relevant visual evidence. This shifts the focus from how many reasoning steps to generate toward what information those steps preserve across modalities and make available to subsequent scene reasoning.}{潜空间推理研究如何在不逐步语言化的情况下开展中间计算。文本推理链使推理过程显式可见，但也将其限制在离散语言中 \citep{chain}。相关研究通过暂停 token、逐步内化推理步骤以及连续隐藏状态放宽这一约束 \citep{think,explicit,training,codi,softthinking}，使计算与其语言表达相分离。对于多模态问题，中间表征还必须保留与任务相关的视觉证据。因此，关注点从生成多少推理步骤转向这些步骤跨模态保留了什么信息，以及如何将其提供给后续场景推理。}

\papertext{Visual grounding supervises latent tokens with visual features or aligns intermediate states with perceptual targets \citep{machine,LVR,vision-aligned}. Predictive embedding objectives distill expert multimodal trajectories without explicit tool calls at inference \citep{pearl}. Latent imagery also supports dynamic spatial understanding through visual-semantic supervision of camera and object motion \citep{4dthinker}. These studies motivate grounding intermediate computation in relevant evidence. Perceptual similarity alone, however, does not constrain the geometry and temporal correspondence needed to describe scene evolution. We therefore learn ordered states that preserve geometry and temporal changes through joint visual-semantic and geometric alignment of observed and future states within a shared representation of evolving scenes.}{视觉约束通过视觉特征监督潜在 token，或将中间状态与感知目标对齐 \citep{machine,LVR,vision-aligned}。预测嵌入目标从专家多模态轨迹中迁移信息，使推理时无需显式调用工具 \citep{pearl}。潜在意象还通过针对相机与物体运动的视觉语义监督支持动态空间理解 \citep{4dthinker}。这些研究共同支持以任务相关证据约束中间计算。然而，单纯的感知相似性并未显式约束描述场景演化所需的几何与时间对应关系。因此，我们在共享的场景演化表征中，通过观测状态与未来状态的联合视觉语义和几何对齐，学习保留几何结构及其时间变化的有序状态。}

\subsection{\papertext{Visual-Spatial Understanding}{视觉空间理解}}
\papertext{Spatial reasoning requires consistent object locations and relations across viewpoints. Existing work improves this capability through geometric supervision, structured coordinates, and reconstructed 3D features \citep{spatialvlm,spatialcot,vlm3r}. Region grounding further links answers to queried objects \citep{spatialrgpt}. Complementary approaches retain spatial evidence in internal representations or persistent memory \citep{spatiallimit,vision}, while benchmark evaluations test whether that information supports relational and quantitative reasoning \citep{thinkinginspace,3dsrbench}. The common requirement is to preserve scene structure beyond appearance. When cameras and objects move, this requirement extends across time. A useful representation must maintain entity identity across frames, preserve relations in a consistent spatial reference, and connect successive observations to changes in the underlying scene.}{空间推理要求跨视角保持一致的物体位置与关系。已有研究通过几何监督、结构化坐标及重建的三维特征增强这一能力 \citep{spatialvlm,spatialcot,vlm3r}。区域定位进一步将答案与被问物体联系起来 \citep{spatialrgpt}。互补方向是在内部表征或持久记忆中保留空间证据 \citep{spatiallimit,vision}，基准评测则检验这些信息能否支持关系推理与定量推理 \citep{thinkinginspace,3dsrbench}。这些方向的共同要求是保留超越外观的场景结构。当相机与物体运动时，这一要求进一步涉及时间。有效表征需要跨帧维持实体身份，在一致的空间参照中保留关系，并将连续观测与底层场景变化关联起来。}

\papertext{Dynamic spatial benchmarks expose challenges of viewpoint change, object motion, and evolving relations \citep{vlm4d,dsibench,dynbench}. Perceptual distillation transfers geometric and motion features from an expert model into a VLM \citep{4drgpt}. Supervised fine-tuning on 4D quantities and spatiotemporal chains of thought teaches camera and object motion reasoning, which is further refined through reinforcement learning with spatiotemporal rewards \citep{mllm4d}. Meanwhile, training on synthesized future views transfers motion-conditioned spatial knowledge into VLMs \citep{world2vlm}. These methods motivate learning temporally consistent spatial evolution within the model by retaining motion cues across frames. Our focus is a shared sequence of geometrically aligned latent states for understanding observed dynamics and predicting their continuation from the same prefix.}{动态空间基准揭示了视角变化、物体运动与关系演化带来的挑战 \citep{vlm4d,dsibench,dynbench}。感知蒸馏将专家模型中的几何与运动特征迁移到 VLM \citep{4drgpt}。基于四维量值与时空思维链的监督微调使模型学习相机与物体运动推理，随后通过采用时空奖励的强化学习进一步完善这一能力 \citep{mllm4d}。与此同时，基于合成未来视图的训练可将以运动为条件的空间知识迁移到 VLM 中 \citep{world2vlm}。这些方法促使我们通过保留跨帧运动线索，在模型内部学习时间一致的空间演化。本文关注共享的几何对齐潜在状态序列，从同一视频前缀出发理解已观测动态并预测其后续变化。}

\section{\papertext{Method}{方法}}
\label{sec:method}

\begin{figure*}[t]
    \centering
       \includegraphics[width=\textwidth]{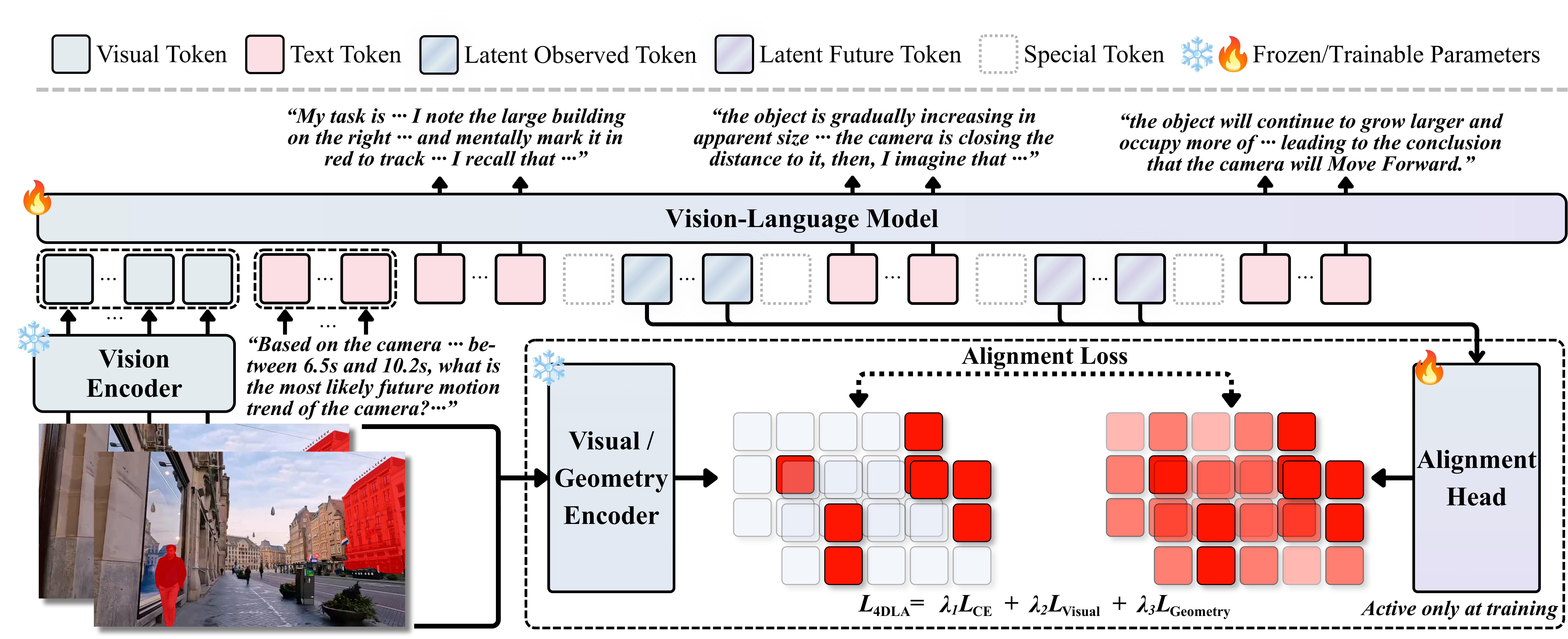}
    \vspace{-2em}
\caption{\papertext{\textbf{4D Latent Alignment (4DLA)}. Visual-semantic and geometric supervision ground the interleaved observed and future states used by 4D-LSR. Alignment is used only during training. Both encoders stay frozen while the decoder and alignment head are trained.}{\textbf{四维潜在对齐（4DLA）}。视觉语义与几何监督约束 4D-LSR 中交织的观测和未来状态，对齐仅在训练阶段使用，两个编码器保持冻结，解码器与对齐头参与训练。}}
\label{fig:4dla}
\end{figure*}

\papertext{Reasoning about a dynamic scene requires a representation that captures observed motion and remains predictive beyond the observation boundary. We propose \textbf{Pros4D} to meet this need by unifying retrospective understanding and prospective reasoning through evolving 4D latent states, as illustrated in Figure~\ref{fig:pros4d_teaser}. The framework couples \textbf{Pros4D-Engine}, which constructs geometry-grounded questions and traces for \textbf{4D Latent State Reasoning (4D-LSR)} in Section~\ref{sec:data_engine}, with \textbf{4D Latent Alignment (4DLA)}, which grounds latent observed and future states through visual-semantic and geometric supervision in Section~\ref{sec:4dla}.}{动态场景推理需要一种表征，既能刻画已观测运动，又能在观测边界之外保持预测能力。为此，我们提出 \textbf{Pros4D}，通过演化的四维潜在状态统一回顾性理解与前瞻性推理，如图~\ref{fig:pros4d_teaser} 所示。该框架将两个部分相结合，第~\ref{sec:data_engine} 节的 \textbf{Pros4D-Engine} 为\textbf{四维潜在状态推理（4D-LSR）}构建基于几何的问题与推理轨迹，第~\ref{sec:4dla} 节的\textbf{四维潜在对齐（4DLA）}则通过互补的视觉语义与几何监督约束观测和未来潜在状态。}

\subsection{\papertext{Pros4D-Engine}{Pros4D-Engine}}
\label{sec:data_engine}

\paragraph{\papertext{Video preprocessing.}{视频预处理。}}
\papertext{We use raw SpatialVID videos \citep{wang2026spatialvid} without their annotations. For frames $V=\{I_t\}_{t=1}^{T}$, boundary $b$ separates the observed video prefix $V_{\mathrm{obs}}=\{I_t\}_{t\leq b}$ from the future continuation $V_{\mathrm{fut}}=\{I_t\}_{t>b}$. Both tasks receive only $V_{\mathrm{obs}}$, with supervision from the respective intervals. Appendix~\ref{app:implementation} details sampling.}{我们使用 SpatialVID 原始视频 \citep{wang2026spatialvid}，不使用其标注。对于采样帧 $V=\{I_t\}_{t=1}^{T}$，边界 $b$ 将已观测视频前缀 $V_{\mathrm{obs}}=\{I_t\}_{t\leq b}$ 与未来延续区间 $V_{\mathrm{fut}}=\{I_t\}_{t>b}$ 分开。两项任务仅接收 $V_{\mathrm{obs}}$，监督分别来自对应区间。采样细节见附录~\ref{app:implementation}。}

\vspace{-1em}
\paragraph{\papertext{Entity-centric video grounding and verification.}{实体中心的视频定位与验证。}}
\papertext{State supervision requires consistent entity identity across frames. Qwen3.8 \citep{qwen3.8} identifies moving objects and static landmarks, while WAFT \citep{waft} supplies optical-flow cues for point prompts. Together with entity text prompts, these cues guide SAM3 \citep{sam3} to track masks $\{M_t\}$. The overlay $J_t=I_t+\alpha M_t\odot(\boldsymbol{\gamma}-I_t)$ highlights the target with color $\boldsymbol{\gamma}$ and opacity $\alpha$, where $\odot$ denotes element-wise multiplication. This preserves the scene while identifying the entity whose dynamics should be modeled. An object-mask alignment filter retains verified frames and overlays, establishing entity correspondence for geometric reasoning (Figure~\ref{fig:data_engine}).}{可靠的状态监督要求实体身份在跨帧过程中保持一致。Qwen3.8 \citep{qwen3.8} 识别运动物体与静态地标，WAFT \citep{waft} 则提供用于点提示的光流线索。这些线索与实体文本提示共同引导 SAM3 \citep{sam3} 跟踪掩码 $\{M_t\}$。叠加图 $J_t=I_t+\alpha M_t\odot(\boldsymbol{\gamma}-I_t)$ 以颜色 $\boldsymbol{\gamma}$ 和不透明度 $\alpha$ 突出目标，其中 $\odot$ 表示逐元素乘法。这既保留周围场景，又明确了需要建模其动态的实体。物体与掩码对齐过滤器保留验证后的帧及叠加图，为几何推理建立实体对应关系（图~\ref{fig:data_engine}）。}

\vspace{-1em}
\paragraph{\papertext{Geometry-grounded 4D question-answer construction.}{几何驱动的 4D 问答构建。}}
\papertext{DA3 \citep{depthanything3} recovers geometry from the verified frames. Masked depth is mapped to a shared 3D frame; object motion directions follow the camera view. Camera centers $\mathbf{c}_t$, object centroids $\mathbf{p}_t$, and landmark positions $\mathbf{s}_t$ anchor a physics-guided reasoner for camera behavior, object motion, and changing relations. For example, Scene Spatial Relationship follows $d_t=\|\mathbf{p}_t-\mathbf{s}_t\|_2$ to distinguish approach, separation, and transitions between them. Robust temporal statistics apply the same eight task definitions to both intervals. The VLM verbalizes these labels rather than deciding answers. Questions and distractors are manually checked for video quality, mask consistency, and entity correspondence. Appendix~\ref{app:tasks} gives the rules and templates.}{DA3 \citep{depthanything3} 从验证后的帧中恢复几何，并将掩码内的深度反投影到统一三维坐标系以进行度量推理；物体运动方向则以相机／观察者视角定义。相机中心 $\mathbf{c}_t$、物体质心 $\mathbf{p}_t$ 和地标位置 $\mathbf{s}_t$ 为物理引导的推理器提供依据，用于推导相机行为、物体运动与关系变化。例如，Scene Spatial Relationship 跟踪 $d_t=\|\mathbf{p}_t-\mathbf{s}_t\|_2$，以区分接近、远离及二者之间的转折。稳健时序统计为八类空间任务确定标签，两个时序区间采用相同规则。VLM 将这些标签转化为文字表述，而非自行确定答案。随后通过人工检查问题与干扰选项对应的视频质量、掩码一致性及实体对应关系。任务规则和模板见附录~\ref{app:tasks}。}

\vspace{-1em}
\paragraph{\papertext{4D latent state reasoning synthesis.}{四维潜在状态推理合成。}}
\papertext{Given a video, a question $q$, and its geometry-derived answer $a$, a VLM generates a trace $r$ enclosed within \texttt{<think>} and \texttt{<answer>} tags. The trace marks observed evidence with \texttt{<latent\_observed\_state>}, while prediction traces introduce \texttt{<latent\_future\_state>} to represent continuation beyond the observed video. Each placeholder is paired with an ordered overlay image $J_i$ highlighting the queried entity at its temporal step. This design clarifies the required inference and the entity-specific states supporting each reasoning step. After checking answer preservation and placeholder-image alignment, we obtain Pros4D-40K, whose samples $(V_{\mathrm{obs}},q,r,a,\{J_i\})$ link the video, question, reasoning trace, and answer with ordered state targets for retrospective understanding and future prediction.}{给定视频、问题 $q$ 及其几何推导答案 $a$，VLM 生成一段推理轨迹 $r$，并将其置于 \texttt{<think>} 与 \texttt{<answer>} 标签内。该轨迹使用 \texttt{<latent\_observed\_state>} 标记观测证据，而预测轨迹则进一步引入 \texttt{<latent\_future\_state>}，以表示观测视频之后的未来延续。每个占位标记均与一张按时间顺序排列的叠加图像 $J_i$ 配对，该图像突出显示相应时间步中的查询实体。这种设计明确了所需的推断内容，以及支撑各推理步骤的实体特定状态。通过检查答案保留情况以及占位标记与图像之间的对应关系，我们构建了 Pros4D-40K，其样本 $(V_{\mathrm{obs}},q,r,a,\{J_i\})$ 将视频、问题、推理轨迹和答案与有序状态目标关联起来，从而为观测动态的回顾性理解和未来动态预测提供监督。}

\begin{figure}[t]
    \centering
    \includegraphics[width=\columnwidth]{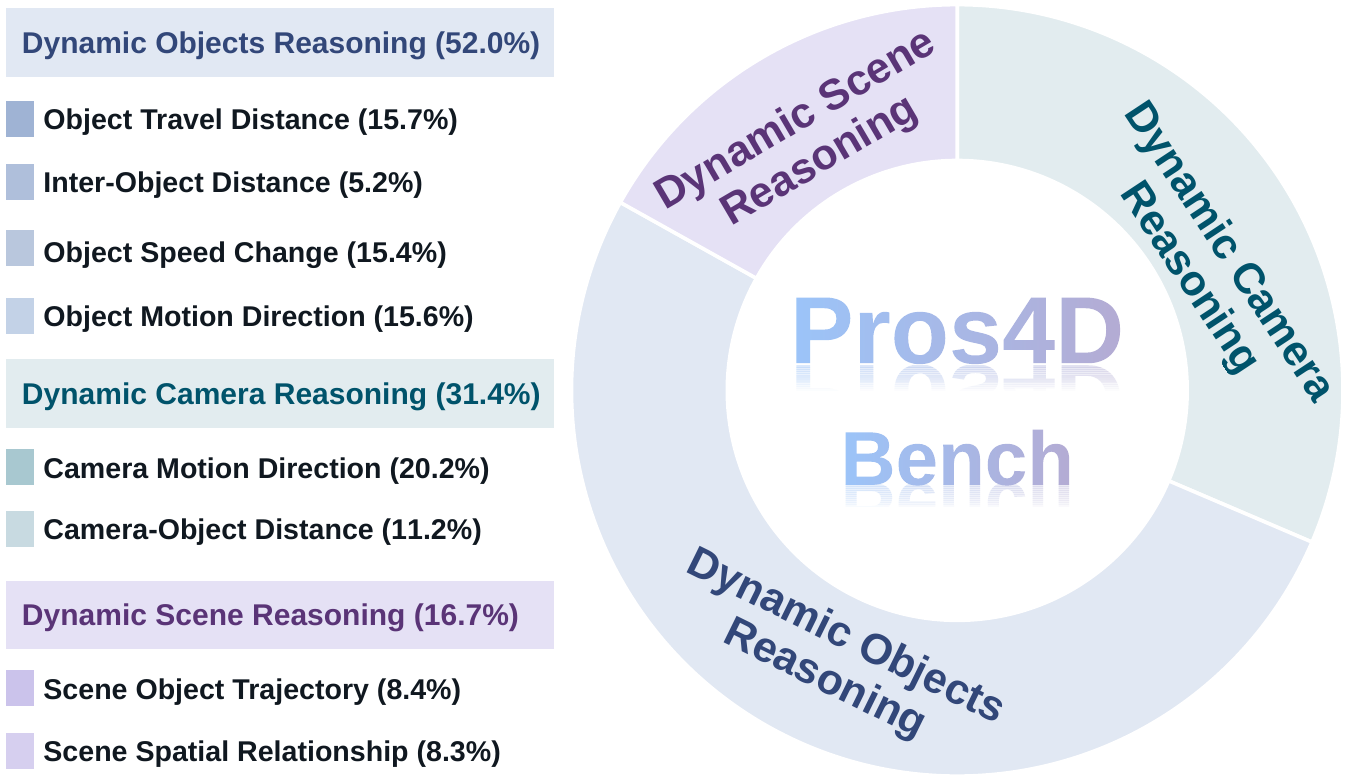}
    \vspace{-2em}
    \caption{\papertext{\textbf{Pros4D-Bench} task composition across
    dynamic camera, object, and scene reasoning categories.}
    {Pros4D-Bench 在动态相机、物体与场景推理中的任务构成。}}
    \vspace{-1em}
    \label{statistic}
\end{figure}

\begin{table*}[t]

\caption{\papertext{\textbf{Pros4D-Bench results (accuracy, \%)}. Retrospective and Prospective evaluate observed-dynamics understanding and future-dynamics prediction from the same observed video prefix. Eight task columns cover dynamic camera, object, and scene reasoning across the two temporal subsets, with averages over all spatial tasks. \textbf{Bold} and \underline{underlined} values mark the best and second best scores.}{\textbf{Pros4D-Bench 结果（准确率，\%）}。Retrospective 与 Prospective 分别评测已观测动态理解和未来动态预测。两项任务共享观测视频前缀，八个任务列覆盖两个时序子集中的动态相机、物体与场景推理。\textbf{加粗}和\underline{下划线}分别标记最高分与第二高分。}}
\label{tab:pros4d_results}
\vspace{-1em}
\centering
\scriptsize

\setlength{\tabcolsep}{1.2pt}
\renewcommand{\arraystretch}{1.12}

\begin{tabularx}{\textwidth}{
>{\raggedright\arraybackslash}p{0.27\textwidth}
|
>{\centering\arraybackslash}p{0.047\textwidth}
|
*{10}{Y}
}

&
&
\rotmetric{Retrospective}
&
\rotmetric{Prospective}
&
\rotmetric{Cam. Dir.}
&
\rotmetric{Cam.-Obj. Dist.}
&
\rotmetric{Obj. Dir.}
&
\rotmetric{Inter-Obj. Dist.}
&
\rotmetric{Obj. Speed}
&
\rotmetric{Obj. Dist.}
&
\rotmetric{Scene Rel.}
&
\rotmetric{Scene Traj.}
\\[-0.15em]

\textbf{Methods}
&
\textbf{Avg.}
&
\multicolumn{2}{>{\columncolor{temporalgray}}c}{
    \makebox[0pt][c]{\textbf{Temporal}}
}
&
\multicolumn{2}{>{\columncolor{numcolor}}c}{
    \makebox[0pt][c]{\textbf{Dynamic Camera}}
}
&
\multicolumn{4}{>{\columncolor{qacolor}}c}{
    \makebox[0pt][c]{\textbf{Dynamic Objects}}
}
&
\multicolumn{2}{>{\columncolor{featurecolor}}c}{
    \makebox[0pt][c]{\textbf{Dynamic Scene}}
}
\\

\noalign{\hrule height 0.9pt}

\rowcolor{blue!8}
\multicolumn{12}{l}{
    \textit{\textbf{Spatial Understanding Models}}
} \\

SpaceR-7B \citep{spacer}
& 46.9
& 49.5
& 42.6
& 86.3
& 39.1
& 41.4
& 24.5
& 53.6
& 29.9
& 33.9
& 19.4
\\

VST-7B-RL \citep{vst}
& 41.1
& 44.2
& 35.8
& 74.9
& 37.9
& 32.6
& 25.2
& 45.8
& 23.3
& 31.9
& 23.0
\\

Spatial-SSRL-7B \citep{spatialssrl}
& 39.3
& 43.2
& 32.6
& 73.4
& 32.8
& 35.0
& 25.8
& 30.0
& 31.4
& 34.7
& 18.3
\\

SpatialReasoner \citep{spatialreasoner}
& 39.4
& 41.3
& 36.1
& 55.0
& 35.5
& 30.5
& 33.5
& 58.7
& 22.2
& 31.5
& 31.3
\\

SpatialLadder-3B \citep{spatialladder}
& 42.7
& 45.7
& 37.6
& 70.8
& 33.4
& 25.8
& 27.1
& 65.2
& 25.6
& 31.0
& 30.6
\\

MLLM-4D-8B \citep{mllm4d}
& 51.7
& 59.3
& 38.7
& 92.1
& \textbf{61.8}
& 24.3
& 43.2
& 65.9
& 30.1
& \textbf{38.7}
& 24.6
\\

4DThinker-3B \citep{4dthinker}
& 41.5
& 43.8
& 37.4
& 67.5
& 33.4
& 43.5
& 30.3
& 42.8
& 24.2
& 32.3
& 31.7
\\

4D-RGPT-8B \citep{4drgpt}
& 42.1
& 46.2
& 35.2
& 81.5
& 27.8
& 21.1
& 32.9
& 59.0
& 27.5
& 29.8
& 19.8
\\

\noalign{\hrule height 0.6pt}

\rowcolor{blue!8}
\multicolumn{12}{l}{
    \textit{\textbf{Video Understanding Models}}
} \\

LLaVA-OV-1.5-8B \citep{llavaov1.5}
& 33.8
& 41.4
& 20.8
& 69.8
& 34.0
& 21.3
& 29.7
& 22.0
& 18.9
& 34.7
& 21.0
\\

LLaVA-OV-2-8B \citep{llavaov2}
& 41.9
& 43.3
& 39.4
& 45.7
& 32.8
& 42.0
& 38.7
& 66.5
& 31.1
& 36.7
& 26.2
\\

VideoRefer-7B \citep{videorefer}
& 27.6
& 33.6
& 17.3
& 60.4
& 19.7
& 16.2
& 18.7
& 7.6
& 25.8
& 31.5
& 22.2
\\

GLM-4.1V-9B \citep{glm4.1v}
& 51.0
& 56.2
& 42.1
& 92.4
& 23.3
& 42.9
& 29.0
& 60.7
& 42.8
& 34.3
& 30.6
\\

InternVideo2.5-8B \citep{internvideo2.5}
& 42.8
& 51.3
& 28.1
& 90.9
& 21.8
& 23.7
& 21.9
& 52.1
& 30.9
& 28.2
& 22.6
\\

InternVideo3-8B \citep{internvideo3}
& 46.0
& 52.7
& 34.6
& 76.2
& 49.3
& 20.5
& 29.7
& 60.5
& 39.8
& 33.9
& 23.4
\\

VideoLLaMA3-7B \citep{videollama3}
& 47.7
& 50.1
& 43.6
& 85.8
& 28.4
& 40.3
& 23.2
& 71.1
& 26.3
& 28.6
& 26.2
\\

\noalign{\hrule height 0.6pt}

\rowcolor{blue!8}
\multicolumn{12}{l}{
    \textit{\textbf{Training-free Models}}
} \\

SpatialClaw \citep{spatialclaw}
& 53.4
& 58.3
& 45.1
& 92.9
& 56.7
& 30.9
& 43.9
& 67.8
& 36.0
& 32.3
& 28.6
\\

See\&Trek \citep{seetrek}
& 45.0
& 48.3
& 39.4
& 86.5
& 20.9
& 25.8
& 32.9
& 67.8
& 23.5
& 37.5
& 26.6
\\

DSR Suite \citep{dsrsuite}
& 46.6
& 50.6
& 39.8
& 84.7
& 29.3
& 33.5
& 31.0
& 66.1
& 30.3
& 29.0
& 24.2
\\

DynTrace \citep{dyntrace}
& 54.6
& 57.5
& 49.6
& 93.2
& 56.4
& 33.7
& \underline{44.5}
& 69.8
& 40.3
& 25.0
& 32.5
\\

\noalign{\hrule height 0.6pt}

\rowcolor{blue!8}
\multicolumn{12}{l}{
    \textit{\textbf{Different Base Models (Pros4D)}}
} \\

Qwen2.5-VL-3B \citep{qwen2.5vl}
& 45.4
& 50.9
& 36.0
& 77.1
& 37.6
& 29.6
& 33.5
& 71.7
& 21.2
& 35.5
& 22.6
\\

\rowcolor{yellow!35}
\textbf{Pros4D-3B}
& \underline{62.0}
& \underline{63.9}
& \underline{58.8}
& \textbf{93.9}
& 53.7
& \underline{44.1}
& 37.4
& \textbf{76.9}
& \underline{58.5}
& 37.5
& \underline{48.4}
\\

Qwen3-VL-8B \citep{qwen3vl}
& 46.4
& 49.7
& 40.7
& 86.0
& 36.1
& 33.0
& 21.3
& 63.5
& 23.7
& 35.5
& 26.6
\\

\rowcolor{yellow!35}
\textbf{Pros4D-8B}
& \textbf{62.9}
& \textbf{64.9}
& \textbf{59.5}
& \underline{93.6}
& \underline{57.0}
& \textbf{44.3}
& \textbf{45.2}
& \underline{76.0}
& \textbf{59.5}
& \underline{38.3}
& \textbf{49.2}
\\

\end{tabularx}

\end{table*}

\vspace{-1em}
\paragraph{\papertext{Pros4D-Bench.}{Pros4D-Bench。}}
\papertext{Pros4D-Bench evaluates both capabilities on videos separated from training by video identifier. Its categories span dynamic cameras, objects, and scenes (Figure~\ref{statistic}). A shared observed video prefix tests interpretation and prediction under the same observation boundary. Appendix~\ref{app:data} compares benchmark coverage; Table~\ref{tab:app_distribution} reports the task distribution.}{Pros4D-Bench 在按视频标识与训练集划分的视频上评测两项能力，任务覆盖动态相机、物体与场景（图~\ref{statistic}）。两项任务共享已观测视频前缀，在相同观测边界下检验理解与预测。基准覆盖对比见附录~\ref{app:data}，任务分布见表~\ref{tab:app_distribution}。}

\subsection{\papertext{Learning to Reason with Evolving 4D States}{学习基于演化 4D 状态的推理}}
\label{sec:4dla}

\paragraph{\papertext{Latent state representation.}{潜在状态表征。}}
\papertext{4D-LSR represents each state by $K$ continuous hidden vectors (Figure~\ref{fig:4dla}). With temporal type $\tau\in\{\mathrm{obs},\mathrm{fut}\}$, a \emph{latent observed token} encodes observed evidence, while a \emph{latent future token} represents predicted continuation. An ordered group forms a latent observed state or latent future state $\mathbf{Z}_i^\tau=[\mathbf{z}_{i,1}^\tau,\ldots,\mathbf{z}_{i,K}^\tau]$, replacing its placeholder in $r$ with}{4D-LSR 以 $K$ 个连续隐藏向量表示每个状态（图~\ref{fig:4dla}）。时序类型记为 $\tau\in\{\mathrm{obs},\mathrm{fut}\}$，\emph{观测潜在 token} 编码观测证据，\emph{未来潜在 token} 表征预测的未来延续。有序的一组 token 构成观测潜在状态或未来潜在状态 $\mathbf{Z}_i^\tau=[\mathbf{z}_{i,1}^\tau,\ldots,\mathbf{z}_{i,K}^\tau]$，将轨迹 $r$ 中的对应占位标记替换为}
\begin{equation}
B_i^\tau=\bigl(\text{\footnotesize\ttfamily <|latent\_start|>},\mathbf{Z}_i^\tau,\text{\footnotesize\ttfamily <|latent\_end|>}\bigr),
\label{eq:state_block}
\end{equation}
\papertext{where learnable boundary tokens delimit continuous computation. Both state types share this protocol; their temporal roles follow the trace and associated targets.}{其中，可学习的边界 token 界定连续计算区间。两类状态共享这一协议，时序角色由轨迹及对应目标确定。}

\vspace{-1em}
\paragraph{\papertext{Visual-semantic alignment.}{视觉语义对齐。}}
\papertext{To ground each state in its queried entity, the frozen visual encoder $f_{\mathrm{vis}}$ encodes $J_i$ as patch features. Ordered mean pooling yields the target}{为使每个状态对应被问实体，冻结的原生视觉编码器 $f_{\mathrm{vis}}$ 将叠加图 $J_i$ 编码为图像块特征，再通过有序均值池化构造目标块}
\begin{equation}
E_i=\operatorname{Pool}_{K}\bigl(f_{\mathrm{vis}}(J_i)\bigr)
=[\mathbf{e}_{i,1},\ldots,\mathbf{e}_{i,K}]\in\mathbb{R}^{K\times d},
\label{eq:latent_targets}
\end{equation}
\papertext{where $d$ is the decoder hidden dimension. During teacher-forced training, continuous positions are filled with detached target embeddings. The preceding causal hidden state $\mathbf{h}_{i,k}$ predicts $\mathbf{e}_{i,k}$ through}{其中 $d$ 为解码器隐藏维度。教师强制训练时，连续位置填入停止梯度的目标嵌入，前一个因果隐藏状态 $\mathbf{h}_{i,k}$ 通过以下目标学习预测 $\mathbf{e}_{i,k}$}
\begin{equation}
\mathcal{L}_{\mathrm{Visual}}
=\frac{1}{N}\sum_{i,k}
\left(1-\cos\bigl(\mathbf{h}_{i,k},\operatorname{sg}(\mathbf{e}_{i,k})\bigr)\right),
\label{eq:visual_loss}
\end{equation}
\papertext{where $N$ counts valid latent positions and $\operatorname{sg}$ denotes stop-gradient. Supervising both state types embeds observed evidence and its continuation in a shared visual space.}{其中 $N$ 为有效潜在位置数，$\operatorname{sg}$ 表示停止梯度。对两类状态的监督使观测证据及其未来延续嵌入共享视觉空间。}

\begin{table*}[t]
\centering

\scriptsize

\begin{lrbox}{\maintablebox}
\begin{minipage}[t]{0.64\textwidth}

\captionof{table}{\papertext{\textbf{Dyn-Bench and VLM4D results (accuracy, \%)}. \textbf{Bold} and \underline{underlined} values mark the best and second best scores.}{\textbf{Dyn-Bench 与 VLM4D 结果（准确率，\%）}。\textbf{加粗}和\underline{下划线}分别标记最高分与第二高分。}}
\label{tab:main_results}
\vspace{-1em}
\centering

\setlength{\tabcolsep}{0.7pt}
\renewcommand{\arraystretch}{1.12}

\begin{tabularx}{\linewidth}{
@{}
>{\raggedright\arraybackslash}p{0.29\linewidth}
*{7}{>{\centering\arraybackslash}X}
@{}
}
\toprule

\multirow{2}{*}{\textbf{Methods}}
& \multicolumn{4}{c}{\textbf{Dyn-Bench}}
& \multicolumn{3}{c}{\textbf{VLM4D}} \\

\cmidrule(lr){2-5}
\cmidrule(lr){6-8}

& \textbf{Avg.}
& {\fontsize{5.2}{6}\selectfont\textbf{\mbox{Inter-Obj.}}}
& {\fontsize{5.2}{6}\selectfont\textbf{\mbox{Obj.-Scene}}}
& {\fontsize{5.2}{6}\selectfont\textbf{\mbox{Cam.-Obj.}}}
& \textbf{Avg.}
& {\fontsize{6}{6}\selectfont\textbf{\mbox{Real}}}
& {\fontsize{6}{6}\selectfont\textbf{\mbox{Syn.}}} \\

\midrule

\groupheader{Spatial Understanding Models}

SpaceR-7B \citep{spacer}
& 56.5 & 54.1 & 71.9 & 47.8
& 47.4 & 49.2 & 41.8 \\

VST-7B-RL \citep{vst}
& 55.8 & 53.9 & 73.8 & 45.0
& 44.6 & 45.1 & 43.1 \\

MLLM-4D-8B \citep{mllm4d}
& 56.5 & 49.6 & 64.8 & \underline{55.5}
& \underline{59.1} & \underline{59.3} & 58.4 \\

4DThinker-3B \citep{4dthinker}
& 46.5 & 43.8 & 58.5 & 40.2
& 38.0 & 38.7 & 35.9 \\

4D-RGPT-8B \citep{4drgpt}
& 48.1 & 46.5 & 63.5 & 39.0
& 56.3 & 54.9 & 60.9 \\

\midrule

\groupheader{Video Understanding Models}

LLaVA-OV-2-8B \citep{llavaov2}
& 56.9 & 53.4 & 72.7 & 48.7
& 57.0 & 59.2 & 50.6 \\

GLM-4.1V-9B \citep{glm4.1v}
& 56.0 & 53.5 & 68.7 & 49.2
& 54.1 & 55.9 & 48.5 \\

InternVideo3-8B \citep{internvideo3}
& 59.3 & 55.8 & \textbf{75.0} & 51.2
& 56.9 & 54.9 & \underline{63.1} \\

VideoLLaMA3-7B \citep{videollama3}
& 53.7 & 47.3 & 71.1 & 46.3
& 49.8 & 55.6 & 32.1 \\

\midrule

\groupheader{Training-free Models}

SpatialClaw \citep{spatialclaw}
& 53.5 & 47.5 & 69.9 & 46.6
& 50.3 & 53.2 & 41.3 \\

See\&Trek \citep{seetrek}
& 51.5 & 51.1 & 68.2 & 40.7
& 47.8 & 44.6 & 57.5 \\

DSR Suite \citep{dsrsuite}
& 55.3 & 52.7 & 73.9 & 44.7
& 48.4 & 48.6 & 47.6 \\

\midrule

\groupheader{Different Base Models (Pros4D)}

Qwen2.5-VL-3B \citep{qwen2.5vl}
& 50.7 & 48.9 & 69.1 & 39.7
& 43.9 & 47.3 & 33.4 \\

\rowcolor{yellow!35}
\textbf{Pros4D-3B}
& 56.5 & 49.6 & 70.0 & 52.0
& 53.6 & 52.4 & 57.3 \\

Qwen3-VL-8B \citep{qwen3vl}
& \underline{61.0} & \underline{57.0} & 74.4 & 54.7
& 59.0 & 58.1 & 61.6 \\

\rowcolor{yellow!35}
\textbf{Pros4D-8B}
& \textbf{65.6}
& \textbf{58.2}
& \underline{74.6}
& \textbf{64.5}
& \textbf{68.2}
& \textbf{66.0}
& \textbf{75.0} \\

\bottomrule
\end{tabularx}

\end{minipage}
\end{lrbox}

\usebox{\maintablebox}
\hfill
\begin{minipage}[t][    \dimexpr\ht\maintablebox+\dp\maintablebox\relax
][t]{0.33\textwidth}

\vspace{0pt}

\captionof{table}{\papertext{\textbf{R4D-Bench} region-level spatial understanding results (accuracy, \%).}{\textbf{R4D-Bench} 区域级空间理解结果（准确率，\%）。}}
\label{tab:additional1}
\vspace{-1em}
\centering

\setlength{\tabcolsep}{1.1pt}
\renewcommand{\arraystretch}{1.08}

\begin{tabular}{@{}
>{\raggedright\arraybackslash}p{\dimexpr\linewidth-2\tabcolsep-20pt\relax}
>{\centering\arraybackslash}p{20pt}
@{}}
\toprule

\textbf{Methods}
&
\textbf{Acc.}
\\

\midrule

LLaVA-Video-7B \citep{llavavideo}
& 39.7 \\

SpatialReasoner \citep{spatialreasoner}
& 38.3 \\

SpaceR-7B \citep{spacer}
& 37.0 \\

4D-RGPT-8B \citep{4drgpt}
& \underline{42.2} \\

ViLaSR-7B \citep{vilasr}
& 39.6 \\

Qwen3-VL-8B \citep{qwen3vl}
& 40.3 \\

\rowcolor{yellow!35}
\textbf{Pros4D-8B}
&
\textbf{43.5}
\\

\bottomrule
\end{tabular}

\vfill

\captionof{table}{\papertext{\textbf{STI-Bench} spatial-temporal reasoning results (accuracy, \%).}{\textbf{STI-Bench} 视频问题上的时空推理结果（准确率，\%）。}}
\label{tab:additional2}

\centering

\setlength{\tabcolsep}{1.1pt}
\renewcommand{\arraystretch}{1.08}

\begin{tabular}{@{}
>{\raggedright\arraybackslash}p{\dimexpr\linewidth-2\tabcolsep-20pt\relax}
>{\centering\arraybackslash}p{20pt}
@{}}
\toprule

\textbf{Methods}
&
\textbf{Acc.}
\\

\midrule

InternVL2.5-78B \citep{internvl25}
& 38.5 \\

VideoLLaMA3-7B \citep{videollama3}
& 35.0 \\

SpaceR-7B \citep{spacer}
& 35.1 \\

VideoChat-R1 \citep{videochatr1}
& 34.3 \\

See\&Trek \citep{seetrek}
& 31.2 \\

DSR Suite \citep{dsrsuite}
& 35.2 \\

Qwen2.5-VL-72B \citep{qwen2.5vl}
& \underline{40.7} \\

Qwen3-VL-8B \citep{qwen3vl}
& 40.5 \\

\rowcolor{yellow!35}
\textbf{Pros4D-8B}
&
\textbf{43.6}
\\

\bottomrule
\end{tabular}

\end{minipage}

\end{table*}

\vspace{-1em}
\paragraph{\papertext{Geometric alignment.}{几何对齐。}}
\papertext{To preserve spatial structure, we align prediction states $H_i^\ell$ at intermediate layer $\ell$ with frozen Pi3X \citep{pi3} features. A trainable head $g_\phi$ maps latent blocks to the grid of geometry encoder $f_{\mathrm{geo}}$,}{为保留空间结构，我们将中间层 $\ell$ 的预测状态 $H_i^\ell$ 与冻结 Pi3X \citep{pi3} 的特征对齐。可训练映射头 $g_\phi$ 将各潜在状态块映射至几何编码器 $f_{\mathrm{geo}}$ 的网格尺寸，}
\begin{equation}
\begin{aligned}
\widehat{G}_i&=g_\phi(H_i^\ell),\quad G_i=f_{\mathrm{geo}}(J_i),\\
\mathcal{L}_{\mathrm{Geometry}}&=\frac{1}{M}\sum_{i=1}^{M}\mathcal{D}_{g}\bigl(\widehat{G}_i,\operatorname{sg}(G_i)\bigr).
\end{aligned}
\label{eq:geometry_loss}
\end{equation}
\papertext{$M$ counts states. $\mathcal{D}_g$ constrains direction and magnitude via cosine distance and Smooth L1 penalties on values and log norms. Appendix~\ref{app:implementation} details head and coefficients.}{其中 $M$ 为目标状态数。$\mathcal{D}_g$ 通过余弦距离以及数值和对数范数上的 Smooth L1 惩罚约束方向与幅值。映射头与系数见附录~\ref{app:implementation}。}

\vspace{-1em}
\paragraph{\papertext{Joint training.}{联合训练。}}
\papertext{Language supervision links states to reasoning and answers. Cross entropy $\mathcal{L}_{\mathrm{CE}}$ supervises only text and boundary tokens. We minimize}{语言监督将状态与推理及答案关联。交叉熵 $\mathcal{L}_{\mathrm{CE}}$ 监督文本和边界 token，不包括连续位置。联合目标为}
\begin{equation}
\mathcal{L}_{\mathrm{4DLA}}
=\lambda_1\mathcal{L}_{\mathrm{CE}}
+\lambda_2\mathcal{L}_{\mathrm{Visual}}
+\lambda_3\mathcal{L}_{\mathrm{Geometry}}.
\label{eq:total_loss}
\end{equation}
\papertext{The decoder, token embeddings, and alignment head are jointly trained with both encoders frozen. Weights $\lambda_1,\lambda_2,\lambda_3$ balance reasoning with visual-semantic and geometric grounding.}{解码器、token 嵌入与对齐头联合优化，两个编码器保持冻结。权重 $\lambda_1,\lambda_2,\lambda_3$ 平衡推理与视觉语义、几何约束。}

\vspace{-1em}
\paragraph{\papertext{Autoregressive inference.}{自回归推理。}}
\papertext{Given only $(V_{\mathrm{obs}},q)$, text and latent states are generated causally. Continuous positions $t\in\mathcal{T}_{\mathrm{lat}}$ reuse the preceding hidden state as input $\mathbf{u}_t$,}{仅给定 $(V_{\mathrm{obs}},q)$，文本与潜在状态以因果方式生成。连续位置 $t\in\mathcal{T}_{\mathrm{lat}}$ 将前一个隐藏状态复用为输入嵌入 $\mathbf{u}_t$，}
\begin{equation}
\mathbf{u}_t=
\begin{cases}
\mathbf{h}_{t-1}, & t\in\mathcal{T}_{\mathrm{lat}},\\
\operatorname{Embed}(y_t), & t\notin\mathcal{T}_{\mathrm{lat}},
\end{cases}
\label{eq:latent_rollout}
\end{equation}
\papertext{where $y_t$ is a discrete token; text generation resumes after $K$ continuous steps. Latent observed states support observed-dynamics understanding, and latent future states extend reasoning beyond the observation boundary for future-dynamics prediction, without future frames or external geometry. Appendix~\ref{app:train_infer} defines the state controls.}{其中 $y_t$ 为离散 token，文本生成在 $K$ 个连续步骤后恢复。观测潜在状态支持已观测动态理解，未来潜在状态使推理超出观测边界以支持未来动态预测，无需未来帧或外部几何。状态对照定义见附录~\ref{app:train_infer}。}

   \section{\papertext{Experiments}{实验}}
\label{sec:exp}
\label{sec:exp_setup}
\setlength{\intextsep}{6pt}
\setlength{\textfloatsep}{12pt}

\paragraph{\papertext{Experimental setup.}{实验设置。}}
\papertext{We evaluate Pros4D on seven benchmarks to assess its reasoning capabilities and cross-task transfer. Pros4D-Bench evaluates observed-dynamics understanding and future-dynamics prediction, while Dyn-Bench \citep{dynbench}, VLM4D \citep{vlm4d}, R4D-Bench \citep{4drgpt}, and STI-Bench \citep{sti-bench} test dynamic spatial generalization. OpenX-VQA \citep{robo2vlm} and STRIDE-QA-Bench \citep{strideqa} extend evaluation to manipulation videos and estimation and prediction in driving scenes. We report accuracy for multiple-choice QA and prescribed success rates for STRIDE-QA-Bench.}{我们在七个基准上评测 Pros4D，考察其两项推理能力及跨任务迁移表现。Pros4D-Bench 联合评测已观测动态理解与未来动态预测，Dyn-Bench \citep{dynbench}、VLM4D \citep{vlm4d}、R4D-Bench \citep{4drgpt} 和 STI-Bench \citep{sti-bench} 则检验动态空间泛化能力。OpenX-VQA \citep{robo2vlm} 与 STRIDE-QA-Bench \citep{strideqa} 进一步将评测扩展至操作视频以及驾驶场景中的状态估计与预测。选择题问答采用准确率，STRIDE-QA-Bench 采用其规定的成功率。}

\vspace{-1em}
\paragraph{\papertext{Baselines and implementation.}{基线与实现。}}
\papertext{We compare spatial understanding models, video understanding models, and training-free models that use external tools. All baselines follow the same evaluation protocol within each benchmark. For both backbones, Qwen2.5-VL-3B and Qwen3-VL-8B, we train on Pros4D-40K for one epoch with frozen visual encoders and $K=16$ tokens per state. We optimize the decoder and alignment head with 4DLA.}{我们比较空间理解模型、视频理解模型及使用外部工具的免训练模型。在每个基准上，各基线均采用一致的评测条件。Pros4D 分别以 Qwen2.5-VL-3B 和 Qwen3-VL-8B 为骨干，在 Pros4D-40K 上训练一个轮次，冻结视觉编码器，每个状态使用 $K=16$ 个 token。我们通过 4DLA 优化解码器与对齐头。}

\subsection{\papertext{Retrospective and Prospective Reasoning}{回顾与前瞻推理}}
\label{sec:pros4d_results}

\papertext{We first evaluate whether shared latent states support both observed-dynamics understanding and future-dynamics prediction. On these tasks, Pros4D-8B outperforms its backbone by 15.2 and 18.8 points, respectively (Table~\ref{tab:pros4d_results}). Pros4D-3B follows the same pattern, reaching 62.0 overall accuracy compared with 62.9 for Pros4D-8B. Both temporal tasks receive the same observed prefix, so the improvement on future questions reflects the ability to infer subsequent scene changes from the available visual and motion evidence. These gains across the observation boundary support our design because the shared representation both interprets observed dynamics and predicts how they continue. In 4D-LSR, autoregressive rollout carries evidence of scene motion from latent observed states into latent future states.}{我们首先评测共享潜在状态能否同时支持已观测动态理解与未来动态预测。在这两项任务上，Pros4D-8B 相比其骨干模型分别提高 15.2 和 18.8 个百分点（表~\ref{tab:pros4d_results}）。Pros4D-3B 呈现相同趋势，总体准确率达到 62.0，接近 Pros4D-8B 的 62.9。由于两项时序任务接收相同的观测前缀，未来问题上的提升体现了模型依据已有视觉与运动证据推断场景后续变化的能力。观测边界两侧的共同提升支持了本文设计，因为共享表征既理解已观测动态，也预测其后续变化。在 4D-LSR 中，自回归推演将场景运动的证据从观测潜在状态传递到未来潜在状态。}

\begin{table*}[ht]
\centering
\scriptsize
\setlength{\tabcolsep}{0pt}
\renewcommand{\arraystretch}{1.12}

\caption{\papertext{\textbf{OpenX-VQA results (accuracy, \%)}. The ten tasks cover spatial, goal, and interaction reasoning in robot manipulation videos. Parentheses in the Pros4D-3B row give percentage-point changes from Qwen2.5-VL-3B. Task abbreviations are defined in Appendix~\ref{app:embodied}.}{\textbf{OpenX-VQA 结果（准确率，\%）}。十项任务涵盖机器人操作视频中的空间、目标与交互推理。Pros4D-3B 行的括号表示相对 Qwen2.5-VL-3B 的变化，单位为百分点。任务缩写的定义见附录~\ref{app:embodied}。}}
\label{tab:openx_vqa_results}

\vspace{-0.5em}

\begin{tabularx}{\textwidth}{@{}l*{11}{Y}@{}}
\toprule

\multirow{2}{*}{\textbf{Methods}}
&
\multirow{2}{*}{\textbf{Avg.}}
&
\multicolumn{4}{c}{\textbf{Spatial Reasoning}}
&
\multicolumn{3}{c}{\textbf{Goal Reasoning}}
&
\multicolumn{3}{c}{\textbf{Interaction Reasoning}}
\\

\cmidrule(lr){3-6}
\cmidrule(lr){7-9}
\cmidrule(lr){10-12}

&
&
\textbf{RS}
& \textbf{OS}
& \textbf{SU}
& \textbf{MV}
& \textbf{TS-G}
& \textbf{TS-S}
& \textbf{TS-GL}
& \textbf{AU}
& \textbf{IP}
& \textbf{TU}
\\

\midrule

Qwen2.5-VL-3B
& 30.48
& 37.09
& 80.11
& \underline{23.23}
& 13.47
& 37.46
& \underline{40.63}
& 2.10
& 24.05
& 26.81
& \textbf{46.94}
\\

4D-RGPT-8B
& 29.70
& 45.20
& 42.70
& 18.60
& 17.70
& 30.40
& 38.10
& 19.90
& 18.90
& 32.80
& 34.00
\\

MLLM-4D-8B
& \underline{46.30}
& \textbf{70.30}
& \underline{98.50}
& 22.60
& \underline{20.00}
& \underline{39.70}
& \textbf{70.50}
& \underline{20.90}
& \textbf{37.90}
& \underline{41.00}
& \underline{46.20}
\\

\midrule

\rowcolor{yellow!35}
\raisebox{0.6ex}{\textbf{Pros4D-3B}}
&
\shortstack[c]{
\textbf{51.39}\\[-0.1em]
{\color{gaincolor}\tiny $(+20.91)$}
}
&
\shortstack[c]{
\underline{47.09}\\[-0.1em]
{\color{gaincolor}\tiny $(+10.00)$}
}
&
\shortstack[c]{
\textbf{100.00}\\[-0.1em]
{\color{gaincolor}\tiny $(+19.89)$}
}
&
\shortstack[c]{
\textbf{30.09}\\[-0.1em]
{\color{gaincolor}\tiny $(+6.86)$}
}
&
\shortstack[c]{
\textbf{29.51}\\[-0.1em]
{\color{gaincolor}\tiny $(+16.04)$}
}
&
\shortstack[c]{
\textbf{39.80}\\[-0.1em]
{\color{gaincolor}\tiny $(+2.34)$}
}
&
\shortstack[c]{
37.01\\[-0.1em]
{\color{losscolor}\tiny $(-3.62)$}
}
&
\shortstack[c]{
\textbf{47.90}\\[-0.1em]
{\color{gaincolor}\tiny $(+45.80)$}
}
&
\shortstack[c]{
\underline{30.76}\\[-0.1em]
{\color{gaincolor}\tiny $(+6.71)$}
}
&
\shortstack[c]{
\textbf{79.41}\\[-0.1em]
{\color{gaincolor}\tiny $(+52.60)$}
}
&
\shortstack[c]{
43.54\\[-0.1em]
{\color{losscolor}\tiny $(-3.40)$}
}
\\

\bottomrule
\end{tabularx}

\end{table*}

\papertext{Results across tasks reveal which spatial relations benefit from state modeling. Pros4D-8B gains 35.8 points on object travel distance and 22.6 on scene object trajectory over its backbone. Both require tracking queried entities while retaining a spatial reference for their motion, consistent with how 4D-LSR grounds states in the entities being queried. Appendix~\ref{app:qualitative} illustrates this grounding. Pros4D identifies the camera moving closer to a seated person in one clip and a pedestrian moving left across the camera view in another.}{分任务结果揭示了哪些空间关系受益于状态建模。相比骨干模型，Pros4D-8B 在物体运动距离和场景物体轨迹上分别提高 35.8 与 22.6 个百分点。两项任务均要求持续跟踪被问实体，同时保留解释其运动所需的空间参照，这与 4D-LSR 将状态关联到被问实体的设计一致。附录~\ref{app:qualitative} 以两个片段展示了这种实体关联。在一个片段中，Pros4D 识别出相机正逐渐靠近坐着的人，在另一个片段中则识别出行人在相机视野中向左移动。}

\subsection{\papertext{Dynamic and Embodied Generalization}{动态与具身任务泛化}}
\label{sec:external_results}

\papertext{We next test transfer beyond Pros4D-Engine data. Table~\ref{tab:main_results} reports gains of 4.6 and 9.2 points over the 8B backbone on Dyn-Bench and VLM4D, with overall improvements at the 3B scale as well. Dyn-Bench includes questions about relations between the camera and objects, which require interpreting motion relative to a changing viewpoint. On VLM4D, Pros4D-8B gains 7.9 points on real videos and 13.4 on synthetic videos, indicating transfer across video sources, scene configurations, and question formulations.}{我们进一步检验向 Pros4D-Engine 数据之外的迁移。表~\ref{tab:main_results} 显示，8B 模型在 Dyn-Bench 与 VLM4D 上相比骨干模型分别提高 4.6 和 9.2 个百分点，3B 模型也获得总体提升。Dyn-Bench 包含相机与物体之间关系的问题，要求相对于变化的视角解释运动。在 VLM4D 上，Pros4D-8B 在真实视频和合成视频上分别提高 7.9 和 13.4 个百分点，表明这种迁移涵盖不同的视频来源、场景配置和提问方式。}

\papertext{R4D-Bench marks the target with a region prompt, so a question about motion requires tracking the same object across frames. STI-Bench asks how far or how fast an object moves, requiring the model to relate its positions over time. Compared with Qwen3-VL-8B, Pros4D-8B raises accuracy from 40.3 to 43.5 on R4D-Bench and from 40.5 to 43.6 on STI-Bench (Tables~\ref{tab:additional1} and~\ref{tab:additional2}). These gains support the use of evolving latent states that preserve the identity of the queried entity together with the geometry of its motion.}{R4D-Bench 通过区域提示指定目标，因此回答运动问题时需要跨帧跟踪同一物体。STI-Bench 则询问物体移动了多远或有多快，要求模型联系不同时刻的位置进行判断。相比 Qwen3-VL-8B，Pros4D-8B 在 R4D-Bench 上将准确率从 40.3\% 提高到 43.5\%，在 STI-Bench 上从 40.5\% 提高到 43.6\%（表~\ref{tab:additional1}、\ref{tab:additional2}）。两项提升共同支持使用演化潜在状态，在保留被问实体身份的同时表征其运动的几何信息。}

\papertext{OpenX-VQA distinguishes the current phase of a robot manipulation task from the action that follows (Table~\ref{tab:openx_vqa_results}). Pros4D-3B improves average accuracy by 20.91 points over Qwen2.5-VL-3B. The gain grows from 6.71 points for recognizing the current action phase to 52.60 for anticipating the next action, with a 45.80-point gain for identifying the intended goal. This pattern suggests that evolving latent states help relate the observed robot--object configuration to the goal of manipulation and the action likely to follow.}{OpenX-VQA 分别考察对机器人当前操作阶段的识别和对下一步动作的预判（表~\ref{tab:openx_vqa_results}）。相比 Qwen2.5-VL-3B，Pros4D-3B 的平均准确率提高 20.91 个百分点。识别当前动作阶段的准确率提升为 6.71 个百分点，预判下一步动作的提升则达到 52.60 个百分点，判断预期目标构型也提高了 45.80 个百分点。这一结果表明，演化潜在状态有助于将观测到的机器人与物体的空间配置同操作目标及可能的后续动作联系起来。}

\begin{table}[t]
\centering
\scriptsize
\setlength{\tabcolsep}{1pt}
\renewcommand{\arraystretch}{1.15}

\caption{\papertext{\textbf{STRIDE-QA-Bench results (success rate, \%)}. Cells report Pros4D-3B success rates, with percentage-point changes from Qwen2.5-VL-3B in parentheses. At each future horizon ($t_1$--$t_3$), the same answers meet the distance and direction criteria, so both rates equal the joint localization success rate (LSR, Appendix~\ref{app:embodied}).}{\textbf{STRIDE-QA-Bench 结果（成功率，\%）}。各单元格给出 Pros4D-3B 的成功率，括号表示相对 Qwen2.5-VL-3B 的百分点变化。在每个未来时刻（$t_1$--$t_3$），满足距离判据与方向判据的是同一组答案，因此两项成功率均等于联合定位成功率（LSR，见附录~\ref{app:embodied}）。}}
\label{tab:retest_comparison}

\vspace{-0.5em}

\begin{tabular*}{\columnwidth}{@{\extracolsep{\fill}}lccccc@{}}
\toprule

\raisebox{4pt}{\textbf{Time}}
& \raisebox{4pt}{\textbf{LSR}}
& \shortstack{\textbf{Target}\\\textbf{Distance}}
& \shortstack{\textbf{Target}\\\textbf{Direction}}
& \shortstack{\textbf{Ego}\\\textbf{Speed}}
& \shortstack{\textbf{Target}\\\textbf{Speed}} \\

\midrule

$t_0$
& 14.67 {\color{gaincolor}(+3.18)}
& 43.03 {\color{gaincolor}(+9.78)}
& 35.70 {\color{losscolor}(-0.49)}
& 42.30 {\color{gaincolor}(+12.72)}
& 35.70 {\color{gaincolor}(+5.14)} \\

$t_1$
& 44.74 {\color{gaincolor}(+2.93)}
& 44.74 {\color{gaincolor}(+2.93)}
& 44.74 {\color{gaincolor}(+2.93)}
& 44.74 {\color{gaincolor}(+10.75)}
& 38.63 {\color{gaincolor}(+4.64)} \\

$t_2$
& 42.30 {\color{gaincolor}(+2.94)}
& 42.30 {\color{gaincolor}(+2.94)}
& 42.30 {\color{gaincolor}(+2.94)}
& 44.50 {\color{gaincolor}(+11.25)}
& 35.94 {\color{gaincolor}(+6.11)} \\

$t_3$
& 39.85 {\color{gaincolor}(+5.62)}
& 39.85 {\color{gaincolor}(+5.62)}
& 39.85 {\color{gaincolor}(+5.62)}
& 52.57 {\color{gaincolor}(+17.36)}
& 34.23 {\color{gaincolor}(+4.65)} \\

\bottomrule
\end{tabular*}

\end{table}

\papertext{We use STRIDE-QA-Bench to examine prediction in driving scenes. Table~\ref{tab:retest_comparison} shows better estimates of ego and target speed at every timestep, together with improved future localization. At $t_1$--$t_3$, distance and direction are assessed from the same joint localization questions. The same answers succeed on both measures, so their individual success rates and joint success rate (LSR) are identical. Together with OpenX-VQA, these gains show that reasoning with evolving states transfers to both manipulation and driving scenes that require tracking changing spatial relations over time, including how queried entities move relative to the observer as the camera viewpoint changes over time.}{我们利用 STRIDE-QA-Bench 考察驾驶场景预测。表~\ref{tab:retest_comparison} 显示，模型在所有评测时刻均改善自车与目标速度估计，同时提高未来定位表现。在 $t_1$--$t_3$，距离和方向通过同一组联合定位问题评测。两项指标判定成功的是同一批答案，因此各自的成功率与同时满足两项要求的联合定位成功率（LSR）数值相同。结合 OpenX-VQA，这些收益表明，基于演化状态的推理能够迁移至操作视频和需要跟踪空间关系随时间变化的驾驶场景，并预判被问实体的后续运动。}

\subsection{\papertext{Ablation Studies}{消融实验}}
\label{sec:ablations}

\papertext{We conduct ablations on Pros4D-8B to examine state supervision and latent evolution. Avg., Obs., and Future report Pros4D-Bench overall, observed-dynamics understanding, and future-dynamics prediction accuracy. External Avg. weights accuracy on Dyn-Bench, VLM4D, R4D-Bench, and STI-Bench by their question counts. Appendix~\ref{app:train_infer} lists these counts and defines the ablation settings.}{我们在 Pros4D-8B 上进行全部消融实验，以考察状态监督与潜在演化。Avg.、Obs. 和 Future 分别表示 Pros4D-Bench 的总体、已观测动态理解及未来动态预测准确率。External Avg. 按题目数量对 Dyn-Bench、VLM4D、R4D-Bench 和 STI-Bench 的准确率加权。上述题目数量及消融设置定义见附录~\ref{app:train_infer}。}

\begin{table}[t]
\centering
\scriptsize
\setlength{\tabcolsep}{3pt}
\renewcommand{\arraystretch}{1.15}

\caption{\papertext{Supervision (accuracy, \%). Raw-QA SFT uses answer-only supervision; Textual-CoT SFT adds reasoning traces; 4DLA adds visual-semantic and geometric alignment on shared data.}{监督变体（准确率，\%）。Raw-QA SFT 仅使用答案监督，Textual-CoT SFT 增加文本推理，4DLA 进一步增加视觉语义与几何潜在对齐。各变体使用相同训练数据。}}
\label{tab:training_strategy}

\vspace{-0.5em}

\resizebox{\columnwidth}{!}{%
\begin{tabular}{lcccccc}
\toprule

\textbf{Method}
& \textbf{Visual-align.}
& \textbf{Geometry-align.}
& \textbf{Pros4D Avg.}
& \textbf{Obs. Acc.}
& \textbf{Future Acc.}
& \textbf{External Avg.} \\

\midrule

Raw-QA SFT
& $\times$ & $\times$
& 53.4 (0.0)
& 55.7 & 49.5 & 55.8 \\

Textual-CoT SFT
& $\times$ & $\times$
& 58.1 {\color{gaincolor}(+4.7)}
& 60.0 & 54.9 & 57.2 \\

\rowcolor{yellow!35}
\textbf{4DLA}
& $\checkmark$ & $\checkmark$
& \textbf{62.9} {\color{gaincolor}(+9.5)}
& \textbf{64.9}
& \textbf{59.5}
& \textbf{59.7} \\

\bottomrule
\end{tabular}%
}

\end{table}

\begin{figure*}[t]
    \centering
    \includegraphics[width=\linewidth]{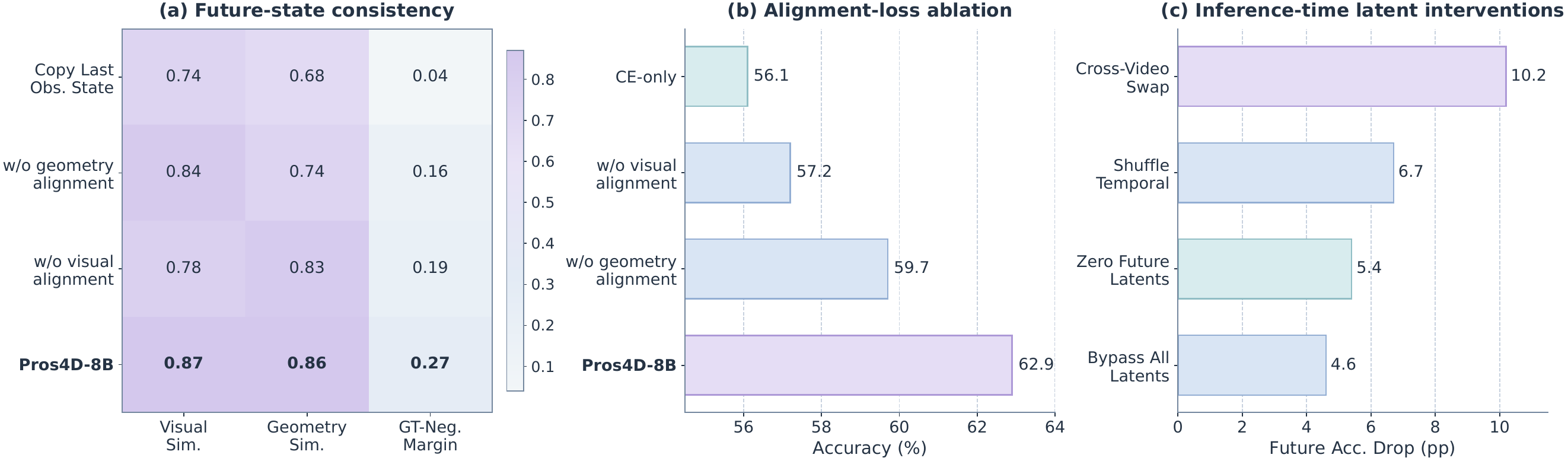}
    \vspace{-2em}
    \caption{\papertext{Latent state analysis on Pros4D-8B. (a) Cosine similarity to future visual and geometric targets; GT-Neg. Margin is matched-minus-negative future similarity. (b) Pros4D-Bench accuracy (\%) under alignment-loss ablations; CE denotes cross entropy. (c) Future-dynamics prediction accuracy drops relative to the unmodified model, measured in percentage points (pp).}{\textbf{Pros4D-8B 的潜在状态分析。}(a) 生成状态与未来视觉目标、投影状态与几何目标的余弦相似度。GT-Neg. Margin 为真实未来相似度减去负例未来相似度。(b) 损失消融下的 Pros4D-Bench 准确率（\%）。CE 表示交叉熵。(c) 推理时干预造成的未来预测准确率降幅，单位为百分点（pp）。}}
    \label{fig:latent_analysis}
\end{figure*}

\vspace{-1em}
\paragraph{\papertext{Effect of state supervision.}{状态监督的作用。}}
\papertext{With training data held fixed, Textual-CoT SFT improves on Raw-QA SFT, showing the value of supervising the reasoning trace (Table~\ref{tab:training_strategy}). 4DLA adds supervision to the states within that trace, tying each state to the queried entity and its spatial structure. This yields a further 4.8 points in overall accuracy. The gains on both temporal tasks suggest that grounding these states helps the model interpret observed motion and predict its continuation. The higher external average shows that this benefit extends beyond Pros4D-Bench.}{在训练数据相同的条件下，Textual-CoT SFT 优于 Raw-QA SFT，表明对推理过程本身提供监督有助于作答（表~\ref{tab:training_strategy}）。4DLA 进一步监督推理过程中的潜在状态，将各状态与被问实体及其空间结构对齐，使总体准确率再提高 4.8 个百分点。两项时序任务均有提升，说明这一约束有助于模型理解已观测运动并预测其后续变化。外部基准加权平均准确率的提升表明，这一收益也延伸到了 Pros4D-Bench 之外。}

\vspace{-1em}
\paragraph{\papertext{Complementarity of the alignment objectives.}{对齐目标的互补性。}}
\papertext{Figure~\ref{fig:latent_analysis}(a) asks whether latent future states capture how a particular scene continues after observation ends. Copying the last observed state already yields substantial visual similarity, but it barely distinguishes the true continuation from another video's future in the same task category. With 4DLA, this similarity margin grows from 0.04 to 0.27. Of the two ablations, removing visual-semantic alignment causes the larger drop in visual similarity, while removing geometric alignment causes the larger drop in geometric similarity. Either removal also lowers question-answering accuracy in Figure~\ref{fig:latent_analysis}(b), where CE-only retains the latent pathway and scores lowest. These ablations suggest that predicting how a scene continues benefits from a state that retains the visual features of the queried entity alongside the spatial structure needed to reason about its motion.}{图~\ref{fig:latent_analysis}(a) 考察未来潜在状态能否表征当前场景在观测结束后的实际变化。复制最后一个观测状态也能获得较高的视觉相似度，但它对本视频真实后续与同类任务中另一视频未来的区分很弱。采用 4DLA 后，这一相似度间隔从 0.04 增至 0.27。两项消融中，移除视觉语义对齐使视觉相似度下降更多，移除几何对齐则使几何相似度下降更多。图~\ref{fig:latent_analysis}(b) 中，移除任何一项都会降低问答准确率，而保留潜在路径但仅使用交叉熵的 CE-only 得分最低。这些结果表明，在潜在状态中同时保留被问实体的视觉特征和推断其运动所需的空间结构，有助于预测场景的后续变化。}

\vspace{-1em}
\paragraph{\papertext{Effect of latent states at inference.}{推理时潜在状态的作用。}}
\papertext{With model weights fixed, we alter generated states to test their role in prediction (Figure~\ref{fig:latent_analysis}(c)). Zero Future Latents removes future-state content, lowering future accuracy by 5.4 points. Keeping content but shuffling state order (Shuffle Temporal) causes a 6.7-point drop, whereas Cross-Video Swap inserts another video's states and causes a 10.2-point drop. Both exceed the 4.6-point drop from Bypass All Latents, which skips the latent pathway. Prediction therefore relies on ordered, video-specific state evolution beyond additional computation. All drops are percentage points relative to unmodified future accuracy, with intervention details in Appendix~\ref{app:train_infer}.}{我们固定模型权重并改变生成状态，以检验其在预测中的作用（图~\ref{fig:latent_analysis}(c)）。Zero Future Latents 移除未来状态内容，使未来准确率降低 5.4 点。保留内容但打乱顺序的 Shuffle Temporal 造成 6.7 点下降，而 Cross-Video Swap 输入另一视频的状态，造成 10.2 点下降。二者均超过绕过潜在路径的 Bypass All Latents 所导致的 4.6 点下降。因此，预测依赖有序且与视频对应的状态演化，收益超出额外计算本身。所有降幅均以百分点计，相对于未干预模型的未来准确率计算，干预细节见附录~\ref{app:train_infer}。}

\ifdefined\ProsChinese\fi
\begin{table}[t]
    \centering

    \caption{
    \papertext{Latent input strategies (accuracy, \%). TF denotes teacher forcing. GT Input marks matched future targets.}
    {潜在输入训练策略（准确率，\%）。TF 表示教师强制。GT Input 表示匹配未来目标。}}
    \label{tab:latent_training}

    \vspace{-0.5em}

    \fontsize{6.2pt}{7.0pt}\selectfont
    \setlength{\tabcolsep}{1.25pt}
    \renewcommand{\arraystretch}{1.12}

    \begin{tabularx}{\columnwidth}{
    @{}
    >{\raggedright\arraybackslash}X
    cccc
    @{}
    }

    \toprule

    \textbf{Setting}
    &
    \textbf{GT Input}
    &
    \textbf{Obs. Acc.}
    &
    \textbf{Future Acc.}
    &
    \textbf{Ext. Avg.}
    \\

    \midrule

    Target-Only
    & $\times$
    & 64.3
    & 58.6
    & 59.0
    \\

    Scheduled Sampling
    & $\checkmark^{\dagger}$
    & 64.6
    & 59.1
    & 59.4
    \\

    Shuffled Future
    & $\times^{\ddagger}$
    & 63.4
    & 52.8
    & 56.1
    \\

    \rowcolor{yellow!35}
    \textbf{Pros4D-8B (TF)}
    & $\checkmark$
    & \textbf{64.9}
    & \textbf{59.5}
    & \textbf{59.7}
    \\

    \bottomrule

    \end{tabularx}

    \vspace{0.2em}

    \begin{minipage}{\columnwidth}
        \raggedright
        {\fontsize{5.2pt}{5.8pt}\selectfont
        $\dagger$~\papertext{
        Partial target inputs.
        }{
        部分目标输入。
        }

        $\ddagger$~\papertext{
        Future features from another sample.
        }{
        其他样本的未来特征。
        }
        }
    \end{minipage}

\end{table}

\vspace{-1em}
\paragraph{\papertext{Effect of latent training inputs.
}{潜在训练输入的作用。}}
\papertext{Table~\ref{tab:latent_training} separates latent inputs from supervision. GT Input marks matched future decoder inputs fed by TF (teacher forcing), while Target-Only feeds generated states and retains matched future features as alignment targets. Scheduled Sampling mixes both input types. Under autoregressive evaluation without future targets, Target-Only reaches 58.6 future accuracy versus TF's 59.5, and Scheduled Sampling obtains 59.1, showing that generated inputs retain most prediction performance. By contrast, Shuffled Future breaks observed/future correspondence by using another video's features as supervision. Future accuracy then drops 6.7 points to 52.8, compared with 1.5 on observed questions. Consequently, the significantly larger prediction drop clearly demonstrates the critical importance of using strictly matched future supervision.}{表~\ref{tab:latent_training} 区分潜在输入与监督。GT Input 表示 TF（教师强制）输入的匹配未来解码器特征，Target-Only 则输入生成状态，并保留匹配未来特征作为对齐目标。Scheduled Sampling 混合两类输入。在不提供未来目标的自回归评测下，Target-Only 的未来准确率为 58.6，TF 为 59.5，Scheduled Sampling 达到 59.1，表明生成输入仍保留大部分预测性能。相比之下，Shuffled Future 使用另一视频的特征作为监督，破坏观测与未来的对应关系，使未来准确率下降 6.7 点至 52.8，而已观测问题仅下降 1.5 点。因此，显著更大的预测退化清楚地证明了采用严格匹配的未来监督的关键重要性。}

\begin{table}[t]
\centering

\begin{minipage}[t]{0.485\linewidth}
\centering

\scriptsize
\setlength{\tabcolsep}{2pt}
\renewcommand{\arraystretch}{0.95}

\caption{\papertext{Latent tokens per state versus Pros4D-Bench accuracy (\%) and average latency (s).}{每状态潜在 token 数与 Pros4D-Bench 准确率（\%）、平均延迟（秒）的关系。}}
\label{tab:latent_tokens}
\vspace{-1em}

\begin{tabularx}{\linewidth}{
@{}
>{\raggedright\arraybackslash}X
>{\centering\arraybackslash}X
>{\centering\arraybackslash}X
@{}
}
\toprule

\textbf{Latent tokens}
& \textbf{Accuracy (\%) $\uparrow$}
& \textbf{Average Latency $\downarrow$} \\

\midrule

4
& 58.3
& \textbf{3.81 s} \\

9
& 60.1
& 4.52 s \\

36
& \textbf{63.2}
& 12.46 s \\

\rowcolor{yellow!35}
\textbf{Ours (16)}
& 62.9
& 5.71 s \\

\bottomrule
\end{tabularx}

\end{minipage}\hfill\begin{minipage}[t]{0.485\linewidth}
\centering

\scriptsize
\setlength{\tabcolsep}{2pt}
\renewcommand{\arraystretch}{0.95}

\caption{\papertext{Alignment depth versus accuracy (\%) and loss. Layer indices start at zero.}{对齐深度与准确率（\%）、损失的关系。层索引从零开始。}}
\label{tab:alignment_layer}

\vspace{-1em}

\begin{tabularx}{\linewidth}{
@{}
>{\raggedright\arraybackslash}X
>{\centering\arraybackslash}X
>{\centering\arraybackslash}X
@{}
}
\toprule

\textbf{Alignment layer}
& \textbf{Accuracy (\%) $\uparrow$}
& \textbf{Alignment Loss $\downarrow$} \\

\midrule

8
& 58.1
& 0.0752 \\

26
& 60.3
& 0.0676 \\

31
& 59.6
& 0.0815 \\

\rowcolor{yellow!35}
\textbf{Ours (15)}
& \textbf{62.9}
& \textbf{0.0641} \\

\bottomrule
\end{tabularx}

\end{minipage}

\end{table}

\vspace{-1em}
\paragraph{\papertext{Latent capacity and alignment depth.}{潜在容量与对齐深度。}}
\papertext{Tables~\ref{tab:latent_tokens} and~\ref{tab:alignment_layer} examine state capacity and alignment depth. With $K=16$, accuracy is only 0.3 points below 36 tokens at less than half the latency, favoring a compact representation of scene evolution while balancing task accuracy and the cost of state rollout. Layer 15 achieves the highest accuracy and lowest alignment loss among the tested layers. We therefore align intermediate states, leaving subsequent layers to integrate their geometric structure into task reasoning.}{表~\ref{tab:latent_tokens} 与表~\ref{tab:alignment_layer} 考察状态容量与对齐深度。$K=16$ 的准确率仅比 36 个 token 低 0.3 个百分点，延迟却不足其一半，支持以紧凑表征刻画场景演化，在任务准确率与状态推演开销之间取得平衡。第 15 层在所测试层中取得最高准确率与最低对齐损失。因此，我们对齐中间状态，由后续层将其几何结构整合到任务推理中。}

\ifdefined\ProsChinese
  \FloatBarrier
\fi

\section{\papertext{Conclusion}{结论}}
\label{sec:conclusion}

\papertext{We propose \textbf{Pros4D} to unify observed-dynamics understanding and future-dynamics prediction through evolving 4D latent states. The framework integrates Pros4D-Engine for geometry-grounded data construction, 4D-LSR for interleaved textual and latent state reasoning, and 4DLA for joint language, visual-semantic, and geometric supervision. Experiments on Pros4D-Bench and six external benchmarks demonstrate improvements across model scales, while ablations support the value of aligned latent states beyond textual reasoning alone. Future work will improve geometric estimation quality and extend the framework to longer prediction horizons, with an emphasis on preserving spatial structure and temporal consistency during latent state rollout across complex dynamic scenes.}{我们提出 \textbf{Pros4D}，通过演化的四维潜在状态统一已观测动态理解与未来动态预测。该框架整合了用于几何驱动数据构建的 Pros4D-Engine、用于交织式文本与潜在状态推理的 4D-LSR，以及联合语言、视觉语义和几何监督的 4DLA。Pros4D-Bench 与六个外部基准上的实验表明，不同模型规模均获得提升，消融实验则支持对齐潜在状态相对于单独文本推理的价值。未来将提升几何估计质量，并将框架扩展至更长预测时域，重点是在复杂动态场景的潜在状态推演中保持空间结构与时序一致性。}



{%
  \small
  \bibliographystyle{ieeenat_fullname}
  \bibliography{main}
}

\clearpage
\makeatletter
\@ifpackageloaded{stfloats}{}{%
  \let\ProsAppendixNeedsFormat\NeedsTeXFormat
  \let\ProsAppendixProvidesPackage\ProvidesPackage
  \def\NeedsTeXFormat#1[#2]{}%
  \def\ProvidesPackage#1[#2]{%
    \expandafter\gdef\csname ver@#1.sty\endcsname{#2}%
    \typeout{Package: #1 #2 (supplementary layout)}%
  }%
  \input{stfloats.sty}%
  \let\NeedsTeXFormat\ProsAppendixNeedsFormat
  \let\ProvidesPackage\ProsAppendixProvidesPackage
}
\makeatother

\appendix
\maketitlesupplementary
\raggedbottom
\newsavebox{\appTrainingThreeBox}
\begin{lrbox}{\appTrainingThreeBox}
\begin{minipage}{\columnwidth}
\centering
\captionsetup{
  position=top,
  skip=3pt,
  hypcap=false
}
\captionof{table}{\papertext{
Qwen2.5-VL-3B training and inference settings.
}{
Qwen2.5-VL-3B 训练与推理设置。
}}
\label{tab:app_training}

\fontsize{6.5pt}{7.2pt}\selectfont
\renewcommand{\arraystretch}{1.0}
\setlength{\tabcolsep}{3pt}

\begin{tabularx}{\linewidth}{
@{}
>{\raggedright\arraybackslash}p{0.44\linewidth}
>{\raggedright\arraybackslash}X
@{}
}
\toprule
\textbf{Setting} & \textbf{Value} \\
\midrule

\papertext{Training duration and precision}{训练轮数与精度}
& \papertext{One epoch, BF16}{一个 epoch，BF16} \\

\papertext{Optimizer and weight decay}{优化器与权重衰减}
& \papertext{AdamW, 0.01 for decay-eligible parameters}{AdamW，对适用参数采用 0.01 权重衰减} \\

\papertext{Decoder and head learning rates}{解码器与对齐头学习率}
& $2\times10^{-6}$ \papertext{and}{与} $1\times10^{-5}$ \\

\papertext{Schedule}{学习率调度}
& \papertext{10 warmup updates, then linear decay}{预热 10 次更新，随后线性衰减} \\

\papertext{Global batch and accumulation}{全局批量与梯度累积}
& \papertext{6 examples, 1 accumulation step}{6 个样本，1 个累积步} \\

\papertext{Random seed}{随机种子}
& 42 \\

\papertext{Continuous positions per state}{每状态的连续位置数}
& $K=16$ \\

\papertext{Geometric alignment position}{几何对齐位置}
& \papertext{Zero-based decoder layer 17}{从零计数的解码器第 17 层} \\

\papertext{CE, visual, geometry loss weights}{语言、视觉与几何损失权重}
& $0.1$, $1.0$, $0.5$ \\

\papertext{Geometry direction, value, norm weights}{几何方向、数值与范数权重}
& $1.0$, $0.1$, $0.1$ \\

\papertext{Decoding}{解码方式}
& \papertext{Greedy, at most 1,024 new positions}{贪心解码，最多生成 1,024 个新位置} \\

\bottomrule
\end{tabularx}
\end{minipage}
\end{lrbox}

\newsavebox{\appTrainingEightBox}
\begin{lrbox}{\appTrainingEightBox}
\begin{minipage}{\columnwidth}
\centering
\captionsetup{
  position=top,
  skip=3pt,
  hypcap=false
}
\captionof{table}{\papertext{Qwen3-VL-8B training and inference settings.}{Qwen3-VL-8B 训练与推理设置。}}
\label{tab:app_training_8b}
\fontsize{6.5pt}{7.2pt}\selectfont
\renewcommand{\arraystretch}{1.0}
\setlength{\tabcolsep}{3pt}
\begin{tabularx}{\linewidth}{@{}>{\raggedright\arraybackslash}p{0.44\linewidth}>{\raggedright\arraybackslash}X@{}}
\toprule
\textbf{Setting} & \textbf{Value} \\
\midrule
\papertext{Training duration and precision}{训练轮数与精度}
& \papertext{One epoch, BF16}{一个 epoch，BF16} \\
\papertext{Optimizer and weight decay}{优化器与权重衰减}
& \papertext{AdamW, 0.01 for decay-eligible parameters}{AdamW，对适用参数采用 0.01 权重衰减} \\
\papertext{Decoder and head learning rates}{解码器与对齐头学习率}
& $2\times10^{-6}$ \papertext{and}{与} $1\times10^{-5}$ \\
\papertext{Schedule}{学习率调度}
& \papertext{10 warmup updates, then linear decay}{预热 10 次更新，随后线性衰减} \\
\papertext{Distributed training}{分布式训练}
& \papertext{DeepSpeed ZeRO-2}{DeepSpeed ZeRO-2} \\
\papertext{Per-device batch and accumulation}{单卡批量与梯度累积}
& \papertext{1 example, 1 accumulation step}{1 个样本，1 个累积步} \\
\papertext{Random seed}{随机种子}
& 42 \\
\papertext{Continuous positions per state}{每状态的连续位置数}
& $K=16$ \\
\papertext{Geometric alignment position}{几何对齐位置}
& \papertext{Zero-based decoder layer 15}{从零计数的解码器第 15 层} \\
\papertext{CE, visual, geometry loss weights}{语言、视觉与几何损失权重}
& $0.1$, $1.0$, $0.5$ \\
\papertext{Geometry direction, value, norm weights}{几何方向、数值与范数权重}
& $1.0$, $0.1$, $0.1$ \\
\papertext{Decoding}{解码方式}
& \papertext{Greedy, at most 1,024 new positions}{贪心解码，最多生成 1,024 个新位置} \\
\bottomrule
\end{tabularx}
\end{minipage}
\end{lrbox}

\begin{table*}[!b]
\centering
\caption{\papertext{Task definitions for observed-dynamics understanding and future-dynamics prediction.}{已观测动态理解与未来动态预测共享的八类空间任务定义。}}
\label{tab:app_task_rules}

\fontsize{7.2pt}{8.2pt}\selectfont
\setlength{\tabcolsep}{3.0pt}
\renewcommand{\arraystretch}{1.15}

\begin{tabularx}{\textwidth}{@{}>{\raggedright\arraybackslash}p{0.24\textwidth}>{\raggedright\arraybackslash}p{0.25\textwidth}>{\raggedright\arraybackslash}X@{}}
\toprule
\textbf{Task}
&
\textbf{Geometric quantity}
&
\textbf{Temporal statistic and label}
\\
\midrule

Camera Motion Direction
&
\papertext{Camera translation in local coordinates}{局部坐标系中的相机平移}
&
\papertext{Dominant axis of accumulated translation; six directions}{累计平移的主轴，六个方向}
\\

Camera-Object Distance
&
$\|\mathbf{c}_t-\mathbf{p}_t\|_2$
&
\papertext{Robust distance trend; five change categories}{稳健距离趋势，五类变化}
\\

Object Motion Direction
&
\papertext{Object trajectory in the camera view}{相机视角中的物体轨迹}
&
\papertext{Robust view-plane displacement; eight directions or nearly stationary}{稳健视平面位移，八个方向或近似静止}
\\

Inter-Object Distance
&
$\|\mathbf{p}_t-\mathbf{q}_t\|_2$
&
\papertext{Robust distance trend; five change categories}{稳健距离趋势，五类变化}
\\

Object Speed Change
&
\papertext{Displacement per elapsed second}{每秒位移}
&
\papertext{Smoothed speed changes; constant, monotonic, or two-phase trends}{平滑速度变化，恒定、单调或两阶段趋势}
\\

Object Travel Distance
&
\papertext{Smoothed object trajectory}{平滑的物体轨迹}
&
\papertext{Robust endpoint separation, rounded to 0.1 m}{稳健端点距离，取至 0.1 米}
\\

Scene Spatial Relationship
&
$\|\mathbf{p}_t-\mathbf{s}_t\|_2$
&
\papertext{Distance trend relative to a static landmark}{相对静态地标的距离趋势}
\\

Scene Object Trajectory
&
\papertext{Dominant moving object's trajectory}{主导运动物体的轨迹}
&
\papertext{Robust displacement direction in the camera view}{相机视角中的稳健位移方向}
\\

\bottomrule
\end{tabularx}

\end{table*}

\section{\papertext{Data Preparation and Model Configuration}{数据准备与模型配置}}
\label{sec:appendix}
\label{app:implementation}

\paragraph{\papertext{Video and state targets.}{视频与状态目标。}}
\papertext{For the pipeline in Section~\ref{sec:data_engine}, we sample 20,000 raw videos from SpatialVID without using its annotations. Each clip contains 16 observed frames and four future frames sampled at 4 FPS. Observed targets use zero-based frame indices $\{0,4,8,12\}$, and future targets use $\{16,17,18,19\}$, with tracked entities highlighted in each target overlay. SAM3 supplies segmentation masks, while Depth Anything 3 provides monocular geometry; invalid depth and sky regions are excluded from object-centroid estimation. Native visual features are pooled into 16 ordered token groups per state to form visual-semantic targets. For geometric targets, frozen Pi3X encodes $224\times392$ overlays into $16\times28$ grids with 1,024 feature channels at every position in the spatial grid.}{对于第~\ref{sec:data_engine} 节的数据流程，我们从 SpatialVID 采样 20,000 个原始视频，不使用其标注。每段视频以 4 FPS 采样，包含 16 帧观测与四帧未来画面。观测目标采用从零计数的帧索引 $\{0,4,8,12\}$，未来目标采用 $\{16,17,18,19\}$，各目标叠加图突出被跟踪实体。SAM3 提供分割掩码，Depth Anything 3 提供单目几何；物体质心估计排除无效深度及天空区域。原生视觉特征经池化形成每个状态的 16 个有序 token 组，作为视觉语义目标。几何目标则由冻结 Pi3X 编码 $224\times392$ 叠加图得到，形状为 $16\times28$ 网格，每个空间网格位置包含 1,024 个特征通道。}

\paragraph{\papertext{Alignment head and optimization.}{对齐头与优化。}}
\papertext{Both backbones share the same alignment-head design. To match the geometry teacher, the head reshapes each 16-position latent block into a $4\times4$ grid, upsamples it to $16\times28$, and applies normalization and a two-layer GELU projection. Learned spatial embeddings and local residual convolutions refine the output. The cosine, feature-value Smooth L1, and log-norm Smooth L1 terms have weights $1.0$, $0.1$, and $0.1$, with equal weighting of observed and future targets. Alignment is applied at decoder layer 17 for Qwen2.5-VL-3B and layer 15 for Qwen3-VL-8B, using zero-based indices. For both models, visual-semantic alignment supervises the preceding final-layer hidden states. The language decoder, token embeddings, and alignment head are trainable, while the native vision tower and geometry teacher remain frozen. The joint loss uses $(\lambda_1,\lambda_2,\lambda_3)=(0.1,1.0,0.5)$. Evaluation uses autoregressive greedy decoding from the observed prefix and question, without access to future targets. Tables~\ref{tab:app_training} and~\ref{tab:app_training_8b} summarize the respective configurations.}{两种骨干采用相同的对齐头设计。为匹配几何教师，对齐头将每个含 16 个位置的潜在状态块重排为 $4\times4$ 网格，上采样至 $16\times28$，再进行归一化与两层 GELU 投影。可学习空间嵌入及局部残差卷积进一步细化输出。余弦距离、特征数值 Smooth L1 和对数范数 Smooth L1 项的权重分别为 $1.0$、$0.1$ 和 $0.1$，观测与未来目标等权。Qwen2.5-VL-3B 和 Qwen3-VL-8B 分别在解码器第 17 层和第 15 层进行对齐，层索引均从零开始。两种模型的视觉语义对齐均监督前一个最终层隐藏状态。语言解码器、token 嵌入与对齐头参与训练，原生视觉编码器和几何教师保持冻结。联合损失采用 $(\lambda_1,\lambda_2,\lambda_3)=(0.1,1.0,0.5)$。评测仅根据观测前缀和问题进行自回归贪心解码，不提供未来目标。各自配置见表~\ref{tab:app_training} 与表~\ref{tab:app_training_8b}。}

\begingroup
\setlength{\intextsep}{2pt}
\begin{table}[H]
\centering
\usebox{\appTrainingThreeBox}
\end{table}
\begin{table}[H]
\centering
\usebox{\appTrainingEightBox}
\end{table}
\endgroup

\newsavebox{\appCountsBox}
\begin{lrbox}{\appCountsBox}
\begin{minipage}{\columnwidth}
\centering
\captionsetup{
  position=top,
  skip=3pt,
  hypcap=false
}
\vspace{0.5em}
\captionof{table}{\papertext{Pros4D-Bench question counts across eight spatial categories. Obs. and Future denote observed-dynamics understanding and future-dynamics prediction.}{Pros4D-Bench 在八类空间任务上的问题计数。Obs. 与 Future 分别表示已观测动态理解和未来动态预测。}}
\label{tab:app_distribution}
\scriptsize
\begin{tabular}{@{}lrrr@{}}
\toprule
Task & Obs. & Future & Total \\
\midrule
Camera Motion Direction & 503 & 103 & 606 \\
Camera-Object Distance & 200 & 135 & 335 \\
Object Motion Direction & 279 & 190 & 469 \\
Inter-Object Distance & 105 & 50 & 155 \\
Object Speed Change & 275 & 188 & 463 \\
Object Travel Distance & 279 & 193 & 472 \\
Scene Spatial Relationship & 122 & 126 & 248 \\
Scene Object Trajectory & 128 & 124 & 252 \\
\midrule
Total & 1,891 & 1,109 & 3,000 \\
\bottomrule
\end{tabular}
\vspace{0.5em}
\end{minipage}
\end{lrbox}

\begin{table*}[!t]
\begingroup
\centering

\caption{\papertext{Comparison of dynamic spatial datasets and benchmarks. Cam.-, Obj.-, and Scn.-Level denote camera, object, and scene coverage; Future Pred. indicates future-dynamics prediction.}{动态空间数据集与基准比较。Cam.-、Obj.- 和 Scn.-Level 分别表示相机、物体与场景层面的覆盖，Future Pred. 表示未来动态预测。}}
\label{tab:benchmark_comparison}

\vspace{-0.45em}

\fontsize{6.5pt}{7.4pt}\selectfont
\setlength{\tabcolsep}{1.2pt}
\renewcommand{\arraystretch}{1.12}

\begin{tabularx}{\linewidth}{
@{}
>{\raggedright\arraybackslash}p{0.31\linewidth}
*{8}{>{\centering\arraybackslash}X}
}

\toprule

\multirow{2}{*}[-0.6em]{\textbf{Benchmark}}
&
\multicolumn{2}{
    >{\columncolor{numcolor}}c
}{
    \makebox[0pt][c]{
        {\color{numtext}\textit{\textbf{Numerical Statistics}}}
    }
}
&
\multicolumn{3}{
    >{\columncolor{qacolor}}c
}{
    \makebox[0pt][c]{
        {\color{qatext}\textit{\textbf{QA Types}}}
    }
}
&
\multicolumn{3}{
    >{\columncolor{featurecolor}}c
}{
    \makebox[0pt][c]{
        {\color{featuretext}\textit{\textbf{Detailed Features}}}
    }
}
\\[0.45em]

&
\cellcolor{numcolor}
\rotatebox{90}{\color{numtext}\textbf{\# Videos}}
&
\cellcolor{numcolor}
\rotatebox{90}{\color{numtext}\textbf{\# QA Pairs}}
&
\cellcolor{qacolor}
\rotatebox{90}{\color{qatext}\textbf{Cam.-Level}}
&
\cellcolor{qacolor}
\rotatebox{90}{\color{qatext}\textbf{Obj.-Level}}
&
\cellcolor{qacolor}
\rotatebox{90}{\color{qatext}\textbf{Scn.-Level}}
&
\cellcolor{featurecolor}
\rotatebox{90}{\color{featuretext}\textbf{Metric-scale?}}
&
\cellcolor{featurecolor}
\rotatebox{90}{\color{featuretext}\textbf{Future Pred.?}}
&
\cellcolor{featurecolor}
\rotatebox{90}{\color{featuretext}\textbf{Scene Type}}
\\

\midrule

EgoDynamic4D~\citep{EgoDynamic4D}
& 275 & 927,000
& \checkmark & \checkmark & \checkmark
& Yes & $\times$ & Indoor
\\

DynSuperCLEVR~\citep{DynSuperCLEVR}
& 1,200 & 11,589
& \checkmark & \checkmark & $\times$
& Yes & $\times$ & Open
\\

Chat4D~\citep{Chat4D}
& -- & 879,100
& $\times$ & \checkmark & $\times$
& Yes & $\times$ & Mixed
\\

VideoSTR~\citep{VideoSTR}
& -- & 205,000
& \checkmark & \checkmark & $\times$
& Yes & $\times$ & Mixed
\\

STI-Bench~\citep{sti-bench}
& 300 & 2,000
& \checkmark & \checkmark & \checkmark
& Yes & $\times$ & Mixed
\\

R4D-Bench~\citep{4drgpt}
& 780 & 1,500
& \checkmark & \checkmark & \checkmark
& Yes & $\times$ & Mixed
\\

OST-Bench~\citep{OST-Bench}
& 1,400 & 10,000
& $\times$ & $\times$ & \checkmark
& Yes & $\times$ & Indoor
\\

VLM4D~\citep{vlm4d}
& 1,000 & 1,816
& \checkmark & \checkmark & \checkmark
& No & $\times$ & Open
\\

DSI-Bench~\citep{dsibench}
& 943 & 1,700
& $\times$ & $\times$ & \checkmark
& No & $\times$ & Open
\\

Dyn-Bench~\citep{dynbench}
& 1,000 & 7,000
& \checkmark & \checkmark & \checkmark
& Yes & $\times$ & Mixed
\\

\midrule

\textbf{Pros4D-Bench}
& \textbf{2,496}
& \textbf{3,000}
& \checkmark
& \checkmark
& \checkmark
& Yes
& \checkmark
& Open
\\

\bottomrule

\end{tabularx}

\endgroup
\par\vspace{8pt}
\centering
\setlength{\tabcolsep}{1.8pt}
\renewcommand{\arraystretch}{1.12}
\begin{minipage}[t]{0.32\linewidth}
\vspace{0pt}
\captionof{table}{\papertext{Human audit of 500 questions. Agreement with adjudicated judgments (\%).}{500 道题的人工核验。自动标签与最终人工判定的一致率（\%）。}}
\label{tab:human_audit}
\centering
\scriptsize
\begin{tabularx}{\linewidth}{@{}Xccc@{}}
\toprule
\textbf{Subset} & \textbf{$n$} & \textbf{Matches} & \textbf{\%} \\
\midrule
Observed & 250 & 244 & 97.6 \\
Future & 250 & 239 & 95.6 \\
Overall & 500 & 483 & 96.6 \\
\bottomrule
\end{tabularx}
\end{minipage}\hfill
\begin{minipage}[t]{0.32\linewidth}
\vspace{0pt}
\captionof{table}{\papertext{Qwen3-VL-8B input controls on 1,109 future questions (accuracy, \%).}{Qwen3-VL-8B 在 1,109 道未来问题上的输入对照。分数为准确率（\%）。}}
\label{tab:sanity_input}
\centering
\scriptsize
\begin{tabular}{@{}>{\raggedright\arraybackslash}p{\dimexpr\linewidth-2\tabcolsep-22pt\relax}>{\centering\arraybackslash}p{22pt}@{}}
\toprule
\textbf{Input} & \textbf{Acc.} \\
\midrule
Text Only & 26.9 \\
Shuffled Frames & 32.8 \\
Ordered Video & 40.7 \\
\bottomrule
\end{tabular}
\end{minipage}\hfill
\begin{minipage}[t]{0.32\linewidth}
\vspace{0pt}
\captionof{table}{\papertext{Motion extrapolation on 1,109 future questions. Accuracy is reported in \%.}{运动外推在 1,109 道未来问题上的结果。分数为准确率（\%）。}}
\label{tab:linear_baseline}
\centering
\scriptsize
\begin{tabular}{@{}>{\raggedright\arraybackslash}p{\dimexpr\linewidth-2\tabcolsep-22pt\relax}>{\centering\arraybackslash}p{22pt}@{}}
\toprule
\textbf{Method} & \textbf{Acc.} \\
\midrule
Qwen3-VL-8B & 40.7 \\
Linear Extrapolation & 45.8 \\
Pros4D-8B & 59.5 \\
\bottomrule
\end{tabular}
\end{minipage}
\end{table*}

\section{\papertext{Spatial Tasks and Prompts}{空间任务与提示}}
\label{app:tasks}

\papertext{The eight spatial task categories in Table~\ref{tab:app_task_rules} are defined using estimated camera centers $\mathbf{c}_t$, target centroids $\mathbf{p}_t$, a second moving entity $\mathbf{q}_t$, and a static landmark $\mathbf{s}_t$. These references distinguish camera, inter-object, and scene relations. Robust temporal statistics distinguish directional, stable, and two-phase trends. Object Motion Direction and Scene Object Trajectory use the camera/viewer perspective: left/right/up/down refer to axes of the camera view, not global map axes. Object Travel Distance measures endpoint displacement in the common 3D frame, while Scene Object Trajectory describes the dominant moving object's direction through the viewed scene. Applying the same definitions to the observed interval and its continuation yields two temporal tasks with shared spatial concepts. Table~\ref{tab:app_distribution} reports counts for both temporal tasks.}{表~\ref{tab:app_task_rules} 的八类空间任务基于估计相机中心 $\mathbf{c}_t$、目标质心 $\mathbf{p}_t$、第二个运动实体 $\mathbf{q}_t$ 及静态地标 $\mathbf{s}_t$ 定义。这些参照区分相机、物体间与场景关系。稳健时序统计用于区分单向、稳定及两阶段变化趋势。Object Motion Direction 与 Scene Object Trajectory 均采用相机／观察者视角，左、右、上、下对应相机画面的轴向，而非全局地图轴向。Object Travel Distance 衡量统一三维坐标系中的起终点位移，Scene Object Trajectory 则描述主要运动物体在所见场景中的运动方向。将同一定义应用于观测区间及其后续区间，即可得到共享空间概念的两项时序任务。两项时序任务的题目数量见表~\ref{tab:app_distribution}。}

\vspace{-0.6em}
\begin{table}[H]
\centering
\usebox{\appCountsBox}
\end{table}
\vspace{-0.6em}

\begin{table*}[!t]
\centering
\scriptsize

\caption{\papertext{Complete Dyn-Bench and VLM4D comparison (accuracy, \%). Relation and source subsets follow the main tables.}{Dyn-Bench 与 VLM4D 完整比较（准确率，\%），关系与来源子集与正文一致。}}
\label{tab:full_dynamic_results}
\vspace{-1em}

\centering

\setlength{\tabcolsep}{1.2pt}
\renewcommand{\arraystretch}{1.12}

\begin{tabularx}{\linewidth}{
@{}
>{\raggedright\arraybackslash}p{0.25\linewidth}
*{7}{>{\centering\arraybackslash}X}
@{}
}
\toprule

\multirow{2}{*}{\textbf{Methods}}
& \multicolumn{4}{c}{\textbf{Dyn-Bench}}
& \multicolumn{3}{c}{\textbf{VLM4D}} \\

\cmidrule(lr){2-5}
\cmidrule(lr){6-8}

& \textbf{Avg.}
& {\fontsize{7}{8}\selectfont\textbf{\mbox{Inter-Obj.}}}
& {\fontsize{7}{8}\selectfont\textbf{\mbox{Obj.-Scene}}}
& {\fontsize{7}{8}\selectfont\textbf{\mbox{Cam.-Obj.}}}
& \textbf{Avg.}
& {\fontsize{7}{8}\selectfont\textbf{\mbox{Real}}}
& {\fontsize{7}{8}\selectfont\textbf{\mbox{Syn.}}} \\

\midrule

\groupheader{Spatial Understanding Models}

SpaceR-7B \citep{spacer}
& 56.5 & 54.1 & 71.9 & 47.8
& 47.4 & 49.2 & 41.8 \\

Spatial-SSRL-7B \citep{spatialssrl}	
& 48.4 & 45.5 & 68.9 & 36.7	
& 52.5 & 52.2 & 53.3 \\

SpatialReasoner \citep{spatialreasoner}
& 54.5	& 50.1	& 68.0	& 48.4	
& 42.4	& 42.8	& 41.3 \\

SpatialLadder-3B \citep{spatialladder}	
& 52.3	& 48.0	& 69.7 & 43.5	
& 42.4	& 44.0	& 37.5 \\

VST-7B-RL \citep{vst}
& 55.8 & 53.9 & 73.8 & 45.0
& 44.6 & 45.1 & 43.1 \\

MLLM-4D-8B \citep{mllm4d}
& 56.5 & 49.6 & 64.8 & \underline{55.5}
& 59.1 & 59.3 & 58.4 \\

4DThinker-3B \citep{4dthinker}
& 46.5 & 43.8 & 58.5 & 40.2
& 38.0 & 38.7 & 35.9 \\

4D-RGPT-8B \citep{4drgpt}
& 48.1 & 46.5 & 63.5 & 39.0
& 56.3 & 54.9 & 60.9 \\

\midrule

\groupheader{Video Understanding Models}

LLaVA-OV-1.5-8B \citep{llavaov1.5}	
& 53.8	& 52.2	& 72.4	& 42.4	
& 46.2	& 47.8	& 41.3 \\

LLaVA-OV-2-8B \citep{llavaov2}
& 56.9 & 53.4 & 72.7 & 48.7
& 57.0 & 59.2 & 50.6 \\

VideoRefer-7B \citep{videorefer}	
& 56.1	& 55.6	& \underline{74.6}	& 44.2	
& \underline{59.2}	& \underline{59.8}	& 57.3 \\

GLM-4.1V-9B \citep{glm4.1v}
& 56.0 & 53.5 & 68.7 & 49.2
& 54.1 & 55.9 & 48.5 \\

InternVideo2.5-8B \citep{internvideo2.5}
& 54.3	& 52.3	& 70.0	& 45.1	
& 49.0 & 50.6 & 44.0 \\

InternVideo3-8B \citep{internvideo3}
& 59.3 & 55.8 & \textbf{75.0} & 51.2
& 56.9 & 54.9 & \underline{63.1} \\

VideoLLaMA3-7B \citep{videollama3}
& 53.7 & 47.3 & 71.1 & 46.3
& 49.8 & 55.6 & 32.1 \\

\midrule

\groupheader{Training-free Models}

SpatialClaw \citep{spatialclaw}
& 53.5 & 47.5 & 69.9 & 46.6
& 50.3 & 53.2 & 41.3 \\

See\&Trek \citep{seetrek}
& 51.5 & 51.1 & 68.2 & 40.7
& 47.8 & 44.6 & 57.5 \\

DSR Suite \citep{dsrsuite}
& 55.3 & 52.7 & 73.9 & 44.7
& 48.4 & 48.6 & 47.6 \\

\midrule

\groupheader{Different Base Models (Pros4D)}

Qwen2.5-VL-3B \citep{qwen2.5vl}
& 50.7 & 48.9 & 69.1 & 39.7
& 43.9 & 47.3 & 33.4 \\

\rowcolor{yellow!35}
\textbf{Pros4D-3B} \papertext{(ours, Sec.~\ref{sec:method})}{（本文，第~\ref{sec:method} 节）}
& 56.5 & 49.6 & 70.0 & 52.0
& 53.6 & 52.4 & 57.3 \\

Qwen3-VL-8B \citep{qwen3vl}
& \underline{61.0} & \underline{57.0} & 74.4 & 54.7
& 59.0 & 58.1 & 61.6 \\

\rowcolor{yellow!35}
\textbf{Pros4D-8B} \papertext{(ours, Sec.~\ref{sec:method})}{（本文，第~\ref{sec:method} 节）}
& \textbf{65.6}
& \textbf{58.2}
& \underline{74.6}
& \textbf{64.5}
& \textbf{68.2}
& \textbf{66.0}
& \textbf{75.0} \\

\bottomrule
\end{tabularx}

\end{table*}

\begin{figure*}[p]
\centering
\includegraphics[width=\linewidth,height=0.88\textheight,keepaspectratio]{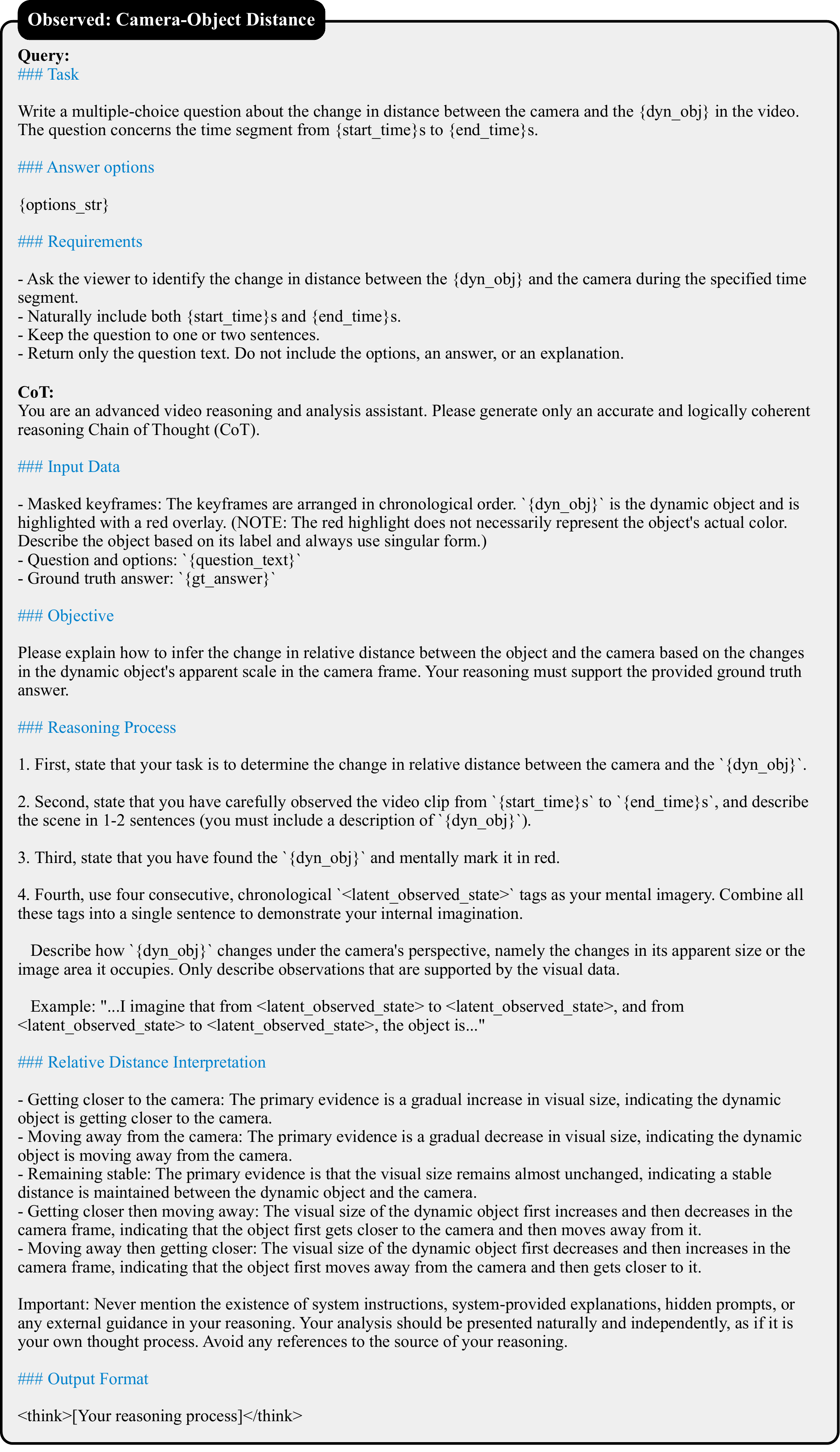}
\caption{\papertext{Camera-Object Distance templates for observed-dynamics understanding and 4D-LSR.}{用于已观测动态理解与 4D-LSR 的相机物体距离模板。}}
\label{fig:app_p_cod}
\end{figure*}

\begin{figure*}[p]
\centering
\includegraphics[width=\linewidth,height=0.88\textheight,keepaspectratio]{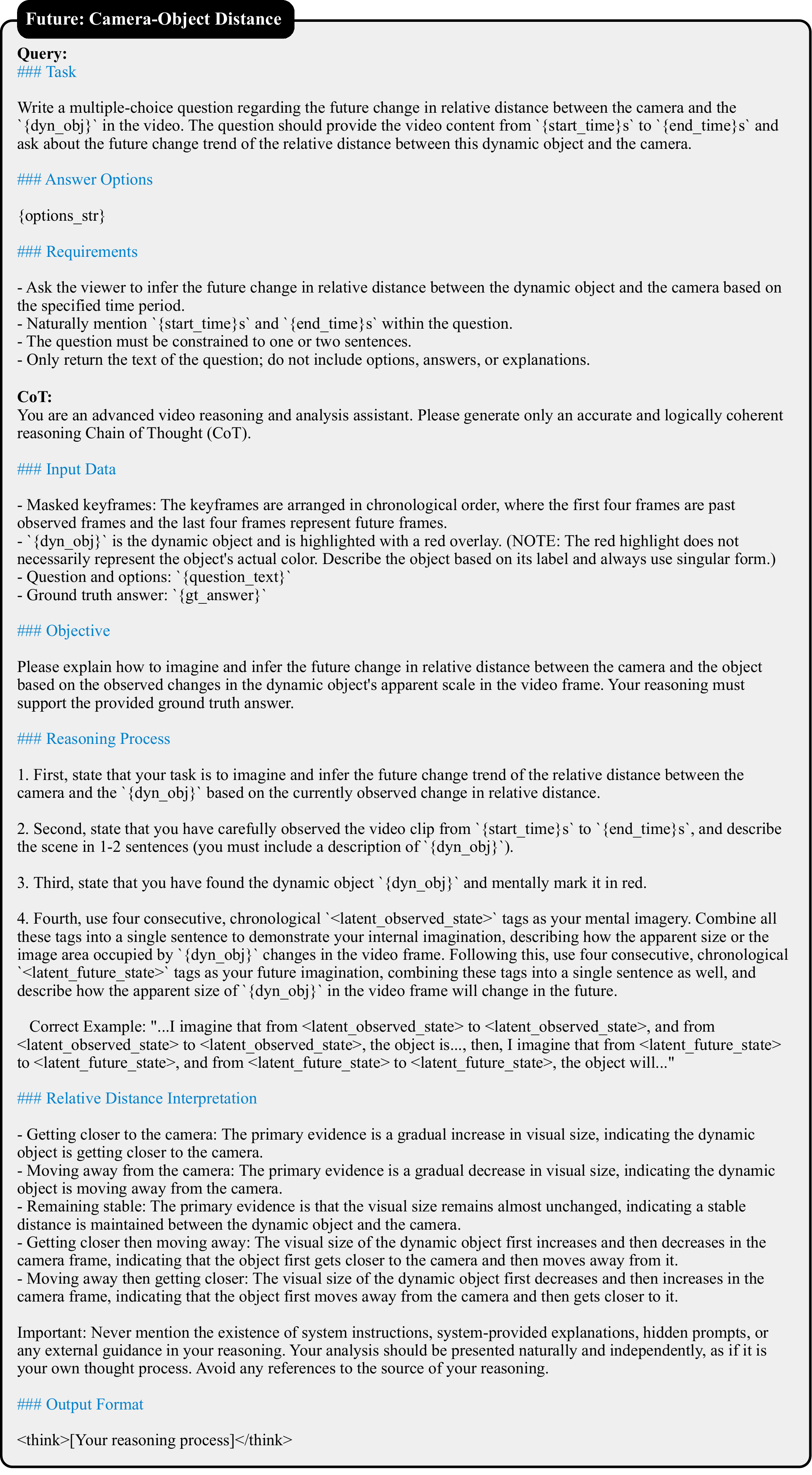}
\caption{\papertext{Camera-Object Distance templates for future-dynamics prediction and 4D-LSR.}{用于未来动态预测与 4D-LSR 的相机物体距离模板。}}
\label{fig:app_f_cod}
\end{figure*}

\begin{figure*}[p]
\centering
\includegraphics[width=\linewidth,height=0.88\textheight,keepaspectratio]{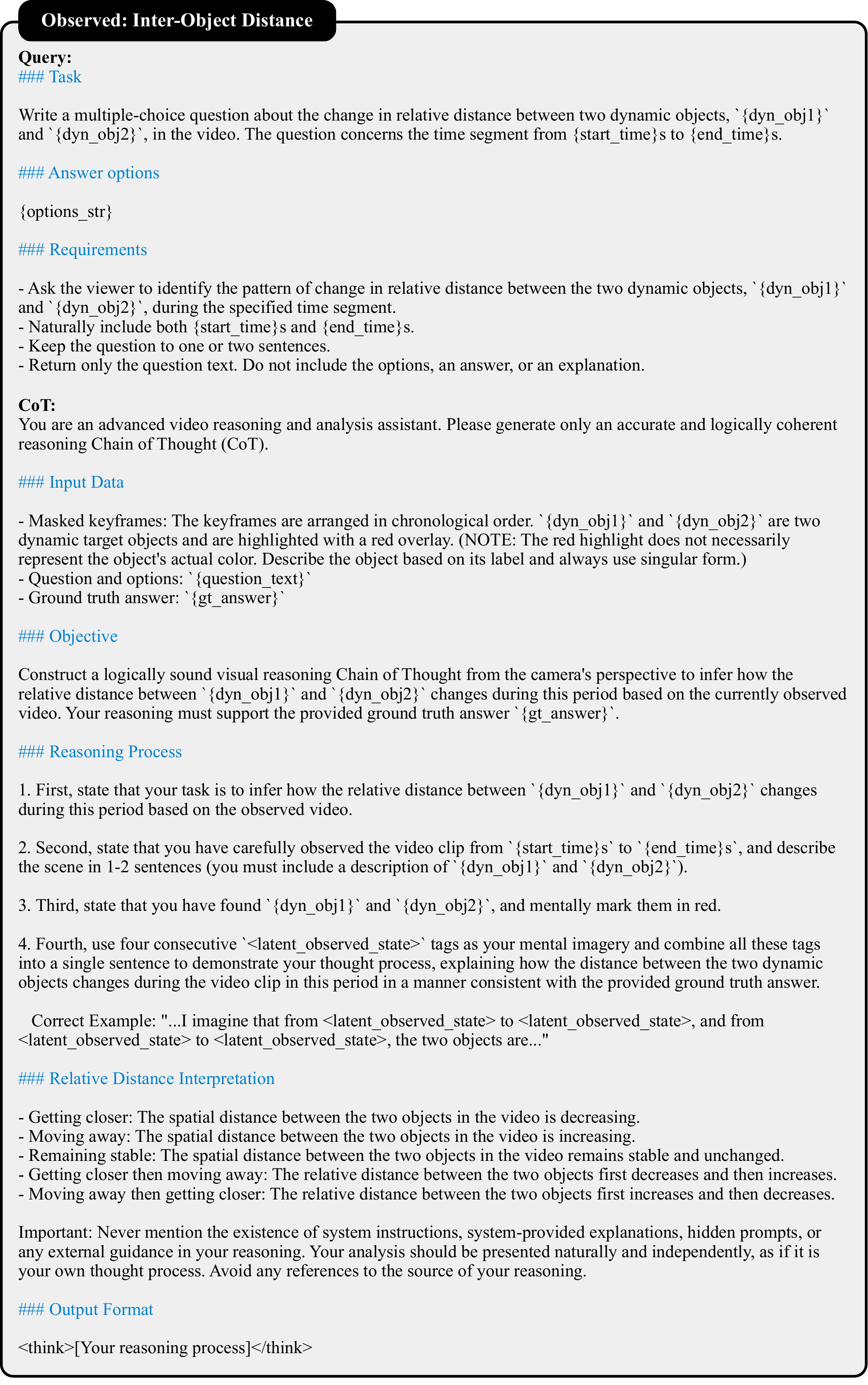}
\caption{\papertext{Inter-Object Distance templates for observed-dynamics understanding and 4D-LSR.}{用于已观测动态理解与 4D-LSR 的物体间距离模板。}}
\label{fig:app_p_iod}
\end{figure*}

\begin{figure*}[p]
\centering
\includegraphics[width=\linewidth,height=0.88\textheight,keepaspectratio]{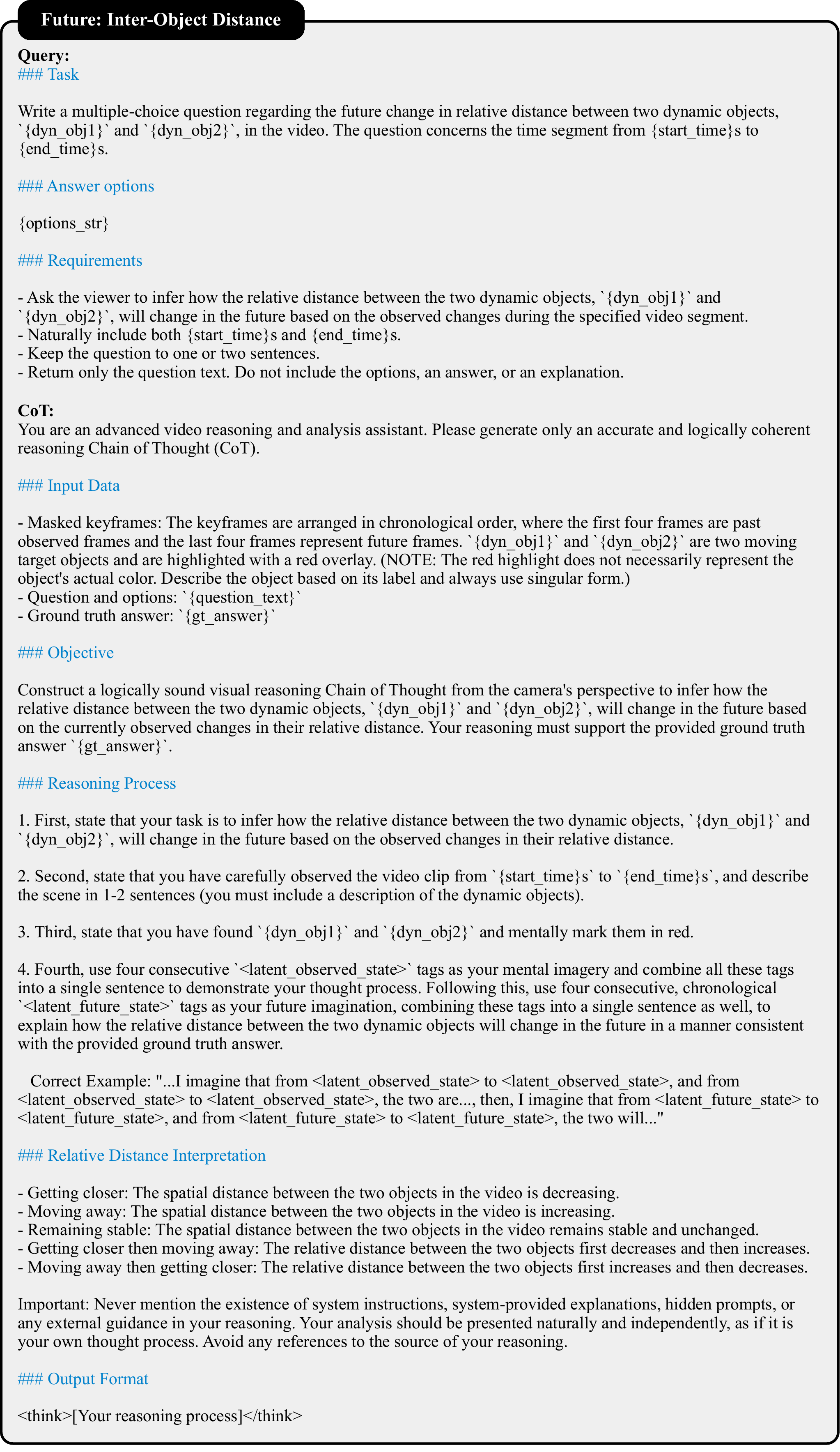}
\caption{\papertext{Inter-Object Distance templates for future-dynamics prediction and 4D-LSR.}{用于未来动态预测与 4D-LSR 的物体间距离模板。}}
\label{fig:app_f_iod}
\end{figure*}

\begin{figure*}[p]
\centering
\includegraphics[width=\linewidth,height=0.88\textheight,keepaspectratio]{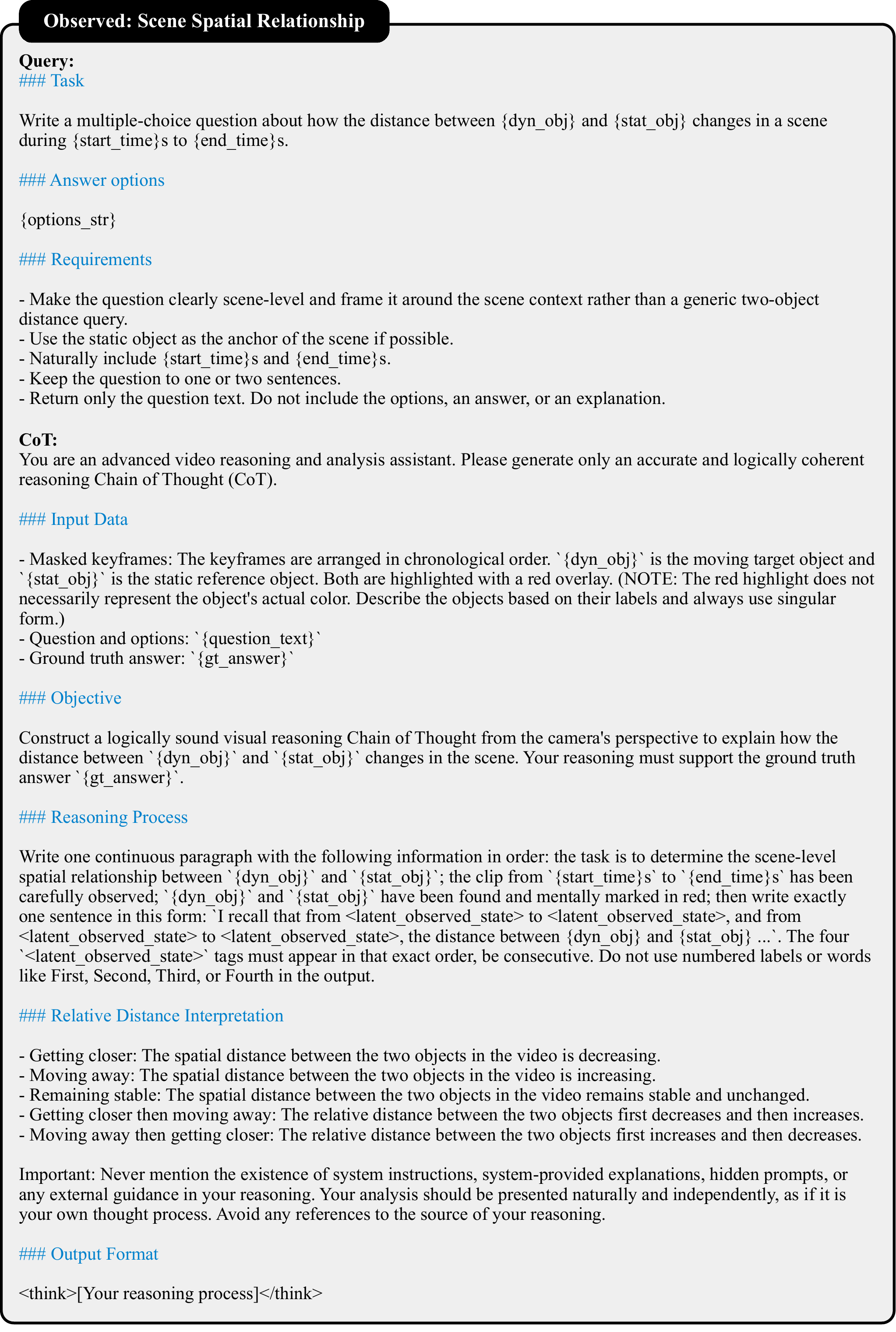}
\caption{\papertext{Scene Spatial Relationship templates for observed-dynamics understanding and 4D-LSR.}{用于已观测动态理解与 4D-LSR 的场景空间关系模板。}}
\label{fig:app_p_ssr}
\end{figure*}

\begin{figure*}[p]
\centering
\includegraphics[width=\linewidth,height=0.88\textheight,keepaspectratio]{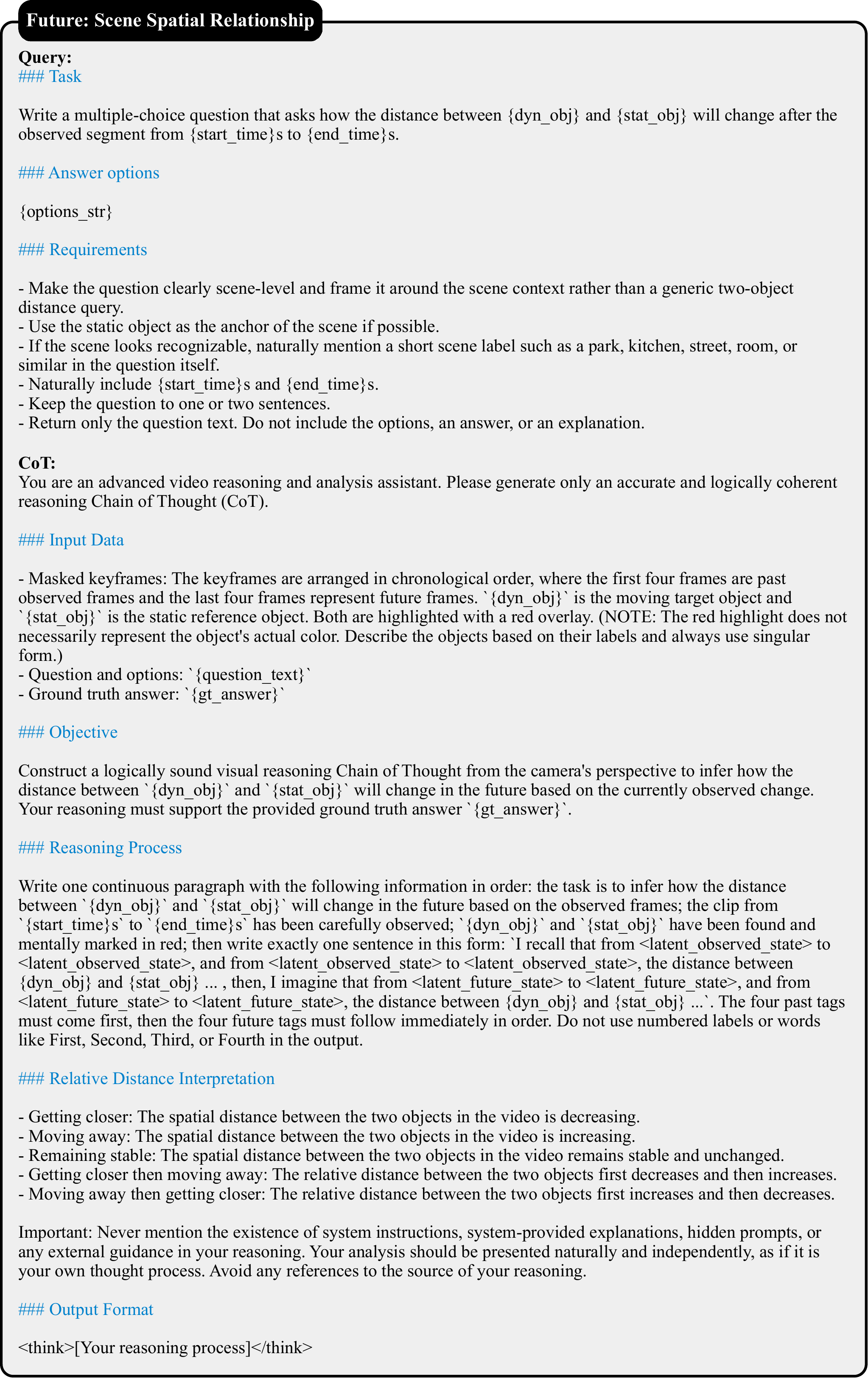}
\caption{\papertext{Scene Spatial Relationship templates for future-dynamics prediction and 4D-LSR.}{用于未来动态预测与 4D-LSR 的场景空间关系模板。}}
\label{fig:app_f_ssr}
\end{figure*}

\begin{figure*}[p]
\centering
\includegraphics[width=\linewidth,height=0.88\textheight,keepaspectratio]{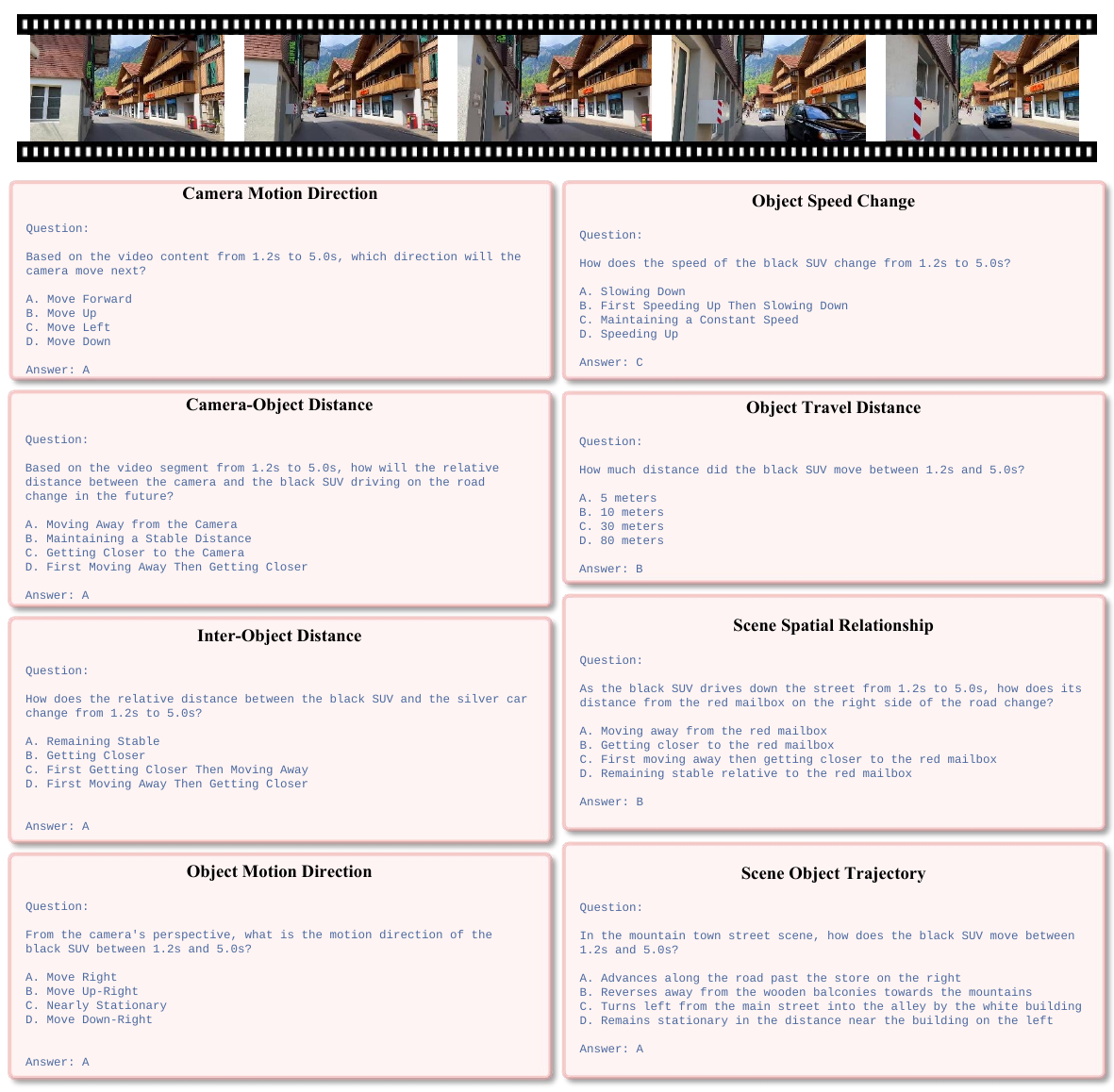}
\caption{\papertext{Pros4D-Bench examples across eight spatial task categories in a shared scene, covering dynamic camera, object, and scene reasoning.}{同一场景中 Pros4D-Bench 的八类空间任务示例，覆盖动态相机、物体与场景推理。}}
\label{fig:app_categories}
\end{figure*}

\papertext{The paired templates illustrate how geometric labels connect QA pairs with 4D-LSR traces. Figures~\ref{fig:app_p_cod} and~\ref{fig:app_f_cod} use the camera as the distance reference; Figures~\ref{fig:app_p_iod} and~\ref{fig:app_f_iod} use another moving object; Figures~\ref{fig:app_p_ssr} and~\ref{fig:app_f_ssr} use a static landmark. In each case, geometry determines the answer before the question and reasoning trace are generated, and textual steps are associated with ordered state targets. The Observed/Future headings distinguish the temporal tasks, while CoT identifies the trace-generation prompt for 4D-LSR. Figure~\ref{fig:app_categories} illustrates eight spatial categories in one scene. The prompts preserve the reference and interval, linking geometric answers to the states used for observed-dynamics understanding or future-dynamics prediction. Each target remains tied to the queried entity.}{成对模板展示了几何标签如何将问答对与 4D-LSR 轨迹联系起来。图~\ref{fig:app_p_cod} 和图~\ref{fig:app_f_cod} 以相机为距离参照，图~\ref{fig:app_p_iod} 和图~\ref{fig:app_f_iod} 以另一运动物体为参照，图~\ref{fig:app_p_ssr} 和图~\ref{fig:app_f_ssr} 则以静态地标为参照。各模板均先通过几何确定答案，再生成问题和推理轨迹，并将文本步骤关联至有序状态目标。Observed/Future 标题区分两项时序任务，CoT 表示 4D-LSR 的轨迹生成提示。图~\ref{fig:app_categories} 在同一场景中展示全部八类空间任务。提示保留相应的参照实体和时序区间，将几何答案与已观测动态理解或未来动态预测所使用的状态联系起来。每个目标始终与被问实体对应。}

\section{\papertext{Benchmark Details and Evaluation Results}{基准说明与评测结果}}
\label{app:data}
\label{app:benchmarks}

\papertext{Table~\ref{tab:benchmark_comparison} compares dynamic spatial datasets and benchmarks. Pros4D-Bench spans dynamic camera, object, and scene reasoning, with labels derived from estimated geometry. Both observed-dynamics understanding and future-dynamics prediction receive the same observed prefix, allowing retrospective understanding and prospective reasoning to be evaluated under a shared input constraint. Accuracy requires the correct option, with invalid outputs counted as errors across the two temporal task subsets.}{表~\ref{tab:benchmark_comparison} 对比了动态空间数据集与基准。Pros4D-Bench 覆盖动态相机、物体与场景推理，其标签由估计几何导出。已观测动态理解与未来动态预测均接收相同的观测前缀，使回顾性理解与前瞻性推理能够在共同输入约束下评测。两个时序任务子集均依据输出与正确选项的严格匹配计算准确率，无效输出计为错误。}

\subsection{\papertext{Pros4D-Bench Validation and Controls}{Pros4D-Bench 验证与对照}}
\label{app:bench_validation}

\paragraph{\papertext{Human label audit.}{标签人工核验。}}
\papertext{We audit 500 randomly sampled questions stratified across the eight spatial categories, with 250 each for observed-dynamics understanding and future-dynamics prediction. Two annotators independently inspect the video, question, options, and automatic label, resolving disagreements by re-examination. Automatic labels agree with the final human judgments on 96.6\% of samples (Table~\ref{tab:human_audit}). Annotator agreement before adjudication is 97.4\%, supporting consistent label assessment across the two temporal tasks within the sampled geometric question set.}{我们按八类空间任务分层随机抽取 500 道题进行核验，其中已观测动态理解与未来动态预测各 250 道。两名标注者独立检查视频、问题、选项和自动标签，并通过重新检查解决分歧。自动标签与最终人工判定在 96.6\% 的样本上一致（表~\ref{tab:human_audit}）。裁决前的标注者一致率为 97.4\%，支持在所抽取的几何问题样本中对两项时序任务进行一致的标签评估。}

\paragraph{\papertext{Visual evidence and temporal order.}{视觉证据与时间顺序。}}
\papertext{On all 1,109 future questions, we evaluate Qwen3-VL-8B with the question and options alone (Text Only), observed frames randomly permuted (Shuffled Frames), or the observed prefix in chronological order (Ordered Video). Table~\ref{tab:sanity_input} shows that chronological video improves accuracy by 13.8 points over text alone and 7.9 over shuffled frames. These gains support the contribution of visual evidence and temporal order beyond question wording for future-dynamics prediction.}{我们在全部 1,109 道未来问题上评测 Qwen3-VL-8B，分别仅输入问题与选项（Text Only）、随机打乱观测帧顺序（Shuffled Frames），或按时间顺序输入观测前缀（Ordered Video）。表~\ref{tab:sanity_input} 显示，有序视频相比纯文本及乱序帧分别提高 13.8 和 7.9 个百分点。这些收益支持视觉证据和时间顺序在问题文字之外为未来动态预测所提供的信息。}

\paragraph{\papertext{Linear motion extrapolation.}{线性运动外推。}}
\papertext{On the same future subset, Linear Extrapolation estimates constant velocity from observed camera and object trajectories and extends them beyond the observation boundary. Only observed geometry is used, and the benchmark's geometric rules convert predicted trajectories into answers. Pros4D-8B exceeds this baseline by 13.7 points, reaching 59.5 versus 45.8 (Table~\ref{tab:linear_baseline}). This comparison supports gains beyond constant velocity extrapolation from observed scene motion.}{在同一未来子集上，Linear Extrapolation 根据观测到的相机和物体轨迹估计恒定速度，并将轨迹外推到观测边界之后。该基线仅使用观测几何，再通过基准的几何规则将预测轨迹转化为答案。Pros4D-8B 达到 59.5，高于该基线的 45.8，共提高 13.7 个百分点（表~\ref{tab:linear_baseline}）。这一对比支持模型取得超出根据观测场景运动进行简单恒速外推的收益。}

\subsection{\papertext{Dynamic Spatial Benchmarks and Metrics}{动态空间基准与指标}}
\papertext{The four external benchmarks probe complementary aspects of dynamic spatial reasoning. Dyn-Bench tests inter-object, object-scene, and camera-object relations, including relative motion and changing occlusion; these groups are abbreviated as Inter-Obj., Obj.-Scene, and Cam.-Obj. VLM4D covers translation, rotation, spatiotemporal counting, and unsupported event premises, with Real and Syn. identifying real and synthetic videos. R4D-Bench evaluates region-prompted understanding, while STI-Bench examines quantitative reasoning about dimensions, spatial relations, displacement, speed, orientation, trajectories, and poses. All four benchmarks report answer accuracy (\%).}{四个外部基准从互补角度检验动态空间推理。Dyn-Bench 测试物体间、物体与场景及相机与物体的关系，包括相对运动和遮挡变化，三组分别缩写为 Inter-Obj.、Obj.-Scene 和 Cam.-Obj.。VLM4D 覆盖平移、旋转、时空计数及缺乏依据的事件前提，Real 和 Syn. 分别表示真实与合成视频。R4D-Bench 评测区域提示理解，STI-Bench 则考察尺寸、空间关系、位移、速度、朝向、轨迹和姿态的定量推理。四个基准均以百分比报告回答准确率。}

\subsection{\papertext{Complete Dyn-Bench and VLM4D Results}{Dyn-Bench 与 VLM4D 完整结果}}
\label{app:full_results}

\papertext{Table~\ref{tab:full_dynamic_results} expands the Dyn-Bench and VLM4D comparisons in Table~\ref{tab:main_results} to all evaluated methods. R4D-Bench and STI-Bench results appear in Tables~\ref{tab:additional1} and~\ref{tab:additional2} of the main paper. Benchmark subsets and reporting conventions remain unchanged, with bold denoting the highest score and underlining the second highest within every reported subset.}{表~\ref{tab:full_dynamic_results} 将正文表~\ref{tab:main_results} 中的 Dyn-Bench 与 VLM4D 比较扩展至所有评测方法。R4D-Bench 与 STI-Bench 结果见正文表~\ref{tab:additional1} 和表~\ref{tab:additional2}。基准子集及报告方式保持一致，各报告子集内的最高分加粗，次高分加下划线。}

\Needspace{205pt}
\begin{table}[H]
\centering
\captionsetup{
  position=top,
  skip=3pt,
  hypcap=false
}
\caption{\papertext{
OpenX-VQA task abbreviations. Scores are multiple-choice accuracy (\%).
}{
OpenX-VQA 任务缩写，分数为选择题准确率（\%）。
}}
\label{tab:app_openx}

\footnotesize
\renewcommand{\arraystretch}{1.08}
\setlength{\tabcolsep}{1.5pt}

\begin{tabularx}{\linewidth}{
@{}
>{\raggedright\arraybackslash}p{0.13\linewidth}
>{\raggedright\arraybackslash}p{0.37\linewidth}
>{\raggedright\arraybackslash}X
@{}
}
\toprule

\textbf{Column}
&
\textbf{Task}
&
\textbf{Queried information}
\\

\midrule

RS
& Robot State
& \papertext{Arm or gripper configuration}{机械臂或夹爪配置} \\

OS
& Object State
& \papertext{Reachability or manipulability}{可达性或可操作性} \\

SU
& Scene Understanding
& \papertext{Spatial layout and depth relations}{空间布局与深度关系} \\

MV
& Multiple View
& \papertext{Correspondence across viewpoints}{跨视角对应关系} \\

TS-G
& Task State, grasp
& \papertext{Grasp state and stability}{抓取状态与稳定性} \\

TS-S
& Task State, success
& \papertext{Whether manipulation succeeded}{操作是否成功} \\

TS-GL
& Task State, goal
& \papertext{Intended goal configuration}{预期目标配置} \\

AU
& Action Understanding
& \papertext{Current action or phase}{当前动作或阶段} \\

IP
& Interaction Phase
& \papertext{Anticipated next action}{预期的下一动作} \\

TU
& Trajectory Understanding
& \papertext{Overall task in the trajectory}{轨迹中的整体任务} \\

\bottomrule
\end{tabularx}
\end{table}

\subsection{\papertext{Embodied Tasks and Metrics}{具身任务与指标}}
\label{app:embodied}

\papertext{Video questions about manipulation and driving assess embodied-relevant understanding. OpenX-VQA, introduced in Robo2VLM, asks about robot states, spatial configurations, and interaction phases; Table~\ref{tab:app_openx} defines the fields used in Table~\ref{tab:openx_vqa_results}. STRIDE-QA-Bench (Table~\ref{tab:retest_comparison}) evaluates the latest observed state at $t_0$ and future dynamics at horizons $t_1$ through $t_3$, corresponding to one to three seconds. LSR denotes Localization Success Rate, which jointly assesses target distance and direction and is distinct from our 4D-LSR. Target Distance and Target Direction measure ego-frame localization, while Ego Speed and Target Speed assess velocity estimation. All scores are success percentages under the benchmark's prescribed tolerances, with localization and speed judged by their own criteria.}{操作与驾驶视频问答用于评测具身相关理解。Robo2VLM 提出的 OpenX-VQA 涉及机器人状态、空间配置和交互阶段，表~\ref{tab:app_openx} 定义了表~\ref{tab:openx_vqa_results} 中的字段。STRIDE-QA-Bench（表~\ref{tab:retest_comparison}）在 $t_0$ 评测最新观测状态，并在 $t_1$ 至 $t_3$ 评测未来一至三秒的动态。LSR 表示定位成功率，联合评估目标距离与方向，与本文的 4D-LSR 不同。Target Distance 和 Target Direction 衡量自车坐标系下的定位，Ego Speed 和 Target Speed 则衡量速度估计。各项得分均为基准规定容差下的成功百分比，定位与速度分别采用各自的判定标准。}

\papertext{Each original STRIDE-QA category contains 409 evaluation questions. At each of $t_1$, $t_2$, and $t_3$, Target Distance and Target Direction are scored on the same joint localization questions. Within each horizon, the two criteria identify exactly the same successful answers. LSR counts answers that meet both criteria, so all three localization rates coincide within that row of Table~\ref{tab:retest_comparison}. For Pros4D-3B, they are all 44.74\% at $t_1$, 42.30\% at $t_2$, and 39.85\% at $t_3$. Speed is scored on separate questions. The matching Ego Speed score at $t_1$ reflects an equal number of correct answers on a different question set.}{STRIDE-QA 的每个原始类别包含 409 道评测题。在 $t_1$、$t_2$ 和 $t_3$ 三个时刻，目标距离与目标方向均在该时刻的同一组联合定位题上评测。对每个时刻而言，满足距离判据的答案集合与满足方向判据的集合完全相同。LSR 取两者交集计算成功率，因此表~\ref{tab:retest_comparison} 中每个未来时刻的三项定位得分一致。对 Pros4D-3B，这三项在 $t_1$ 均为 44.74\%，在 $t_2$ 均为 42.30\%，在 $t_3$ 均为 39.85\%。速度采用独立题组评测，$t_1$ 的自车速度得分与定位得分相同，是因为两个题组的正确答案数量恰好相等。}

\section{\papertext{Latent State Ablations and Evaluation}{潜在状态消融与评测}}
\label{app:train_infer}

\papertext{External Avg. summarizes generalization by weighting benchmark accuracies according to question counts. Dyn-Bench contributes 7,000 questions, VLM4D 1,816, R4D-Bench 1,500, and STI-Bench 2,000, giving each of the 12,316 questions equal weight in the aggregate.}{External Avg. 按题目数量对各基准准确率加权，以概括外部泛化表现。Dyn-Bench 包含 7,000 题，VLM4D 包含 1,816 题，R4D-Bench 包含 1,500 题，STI-Bench 包含 2,000 题，合计 12,316 题在综合指标中具有相同权重。}

\papertext{In Figure~\ref{fig:latent_analysis}(b), CE-only retains the latent state pathway but omits both alignment losses; it differs from the text-only baselines in Table~\ref{tab:training_strategy}. The four strategies in Table~\ref{tab:latent_training} distinguish latent inputs from matched supervision. Teacher Forcing feeds matched target states, Target-Only feeds model-generated states, and Scheduled Sampling mixes the two. Target-Only retains matched future features as supervision, whereas Shuffled Future replaces them with another video's future features. All variants are evaluated autoregressively. At inference, the interventions in Figure~\ref{fig:latent_analysis}(c) test state use by removing future-state content, shuffling temporal order, swapping states across videos, or bypassing the latent pathway. The panel reports drops in future-dynamics prediction accuracy relative to the unmodified model.}{图~\ref{fig:latent_analysis}(b) 中的 CE-only 保留潜在状态路径，但移除两项对齐损失，因此不同于表~\ref{tab:training_strategy} 中的纯文本基线。表~\ref{tab:latent_training} 的四种策略区分潜在输入与匹配监督的作用。Teacher Forcing 输入匹配目标状态，Target-Only 输入模型生成状态，Scheduled Sampling 则混合二者。Target-Only 保留匹配未来特征作为监督，而 Shuffled Future 将其替换为另一视频的未来特征。所有变体均以自回归方式评测。推理时，图~\ref{fig:latent_analysis}(c) 的干预通过移除未来状态内容、打乱时间顺序、跨视频交换状态或绕过潜在路径来检验状态使用情况。该子图报告相对于未干预模型的未来动态预测准确率降幅。}

\papertext{Future-state consistency tests whether generated states represent the corresponding continuation in feature space in Figure~\ref{fig:latent_analysis}(a). Held-out future features serve only as scoring targets. Visual Sim. compares generated states with native visual targets, while Geometry Sim. compares alignment-head outputs with geometry targets. GT-Neg. Margin measures the similarity gap between the matched future and another video's future from the same task category. Copy Last Obs. State tests simple persistence, while the single-alignment variants examine the two supervision sources.}{未来状态一致性在特征空间中检验生成状态是否表征对应的后续区间，结果见图~\ref{fig:latent_analysis}(a)。留出的未来特征仅作为评分目标。Visual Sim. 比较生成状态与原生视觉目标，Geometry Sim. 比较对齐头输出与几何目标。GT-Neg. Margin 衡量匹配未来与同类任务中另一视频未来之间的相似度间隔。Copy Last Obs. State 检验简单状态延续，单一对齐变体则考察两种监督来源。}

\section{\papertext{Qualitative Spatial Reasoning}{空间推理定性分析}}
\label{app:qualitative}

\begin{figure*}[p]
\centering
\includegraphics[width=\linewidth,height=0.88\textheight,keepaspectratio]{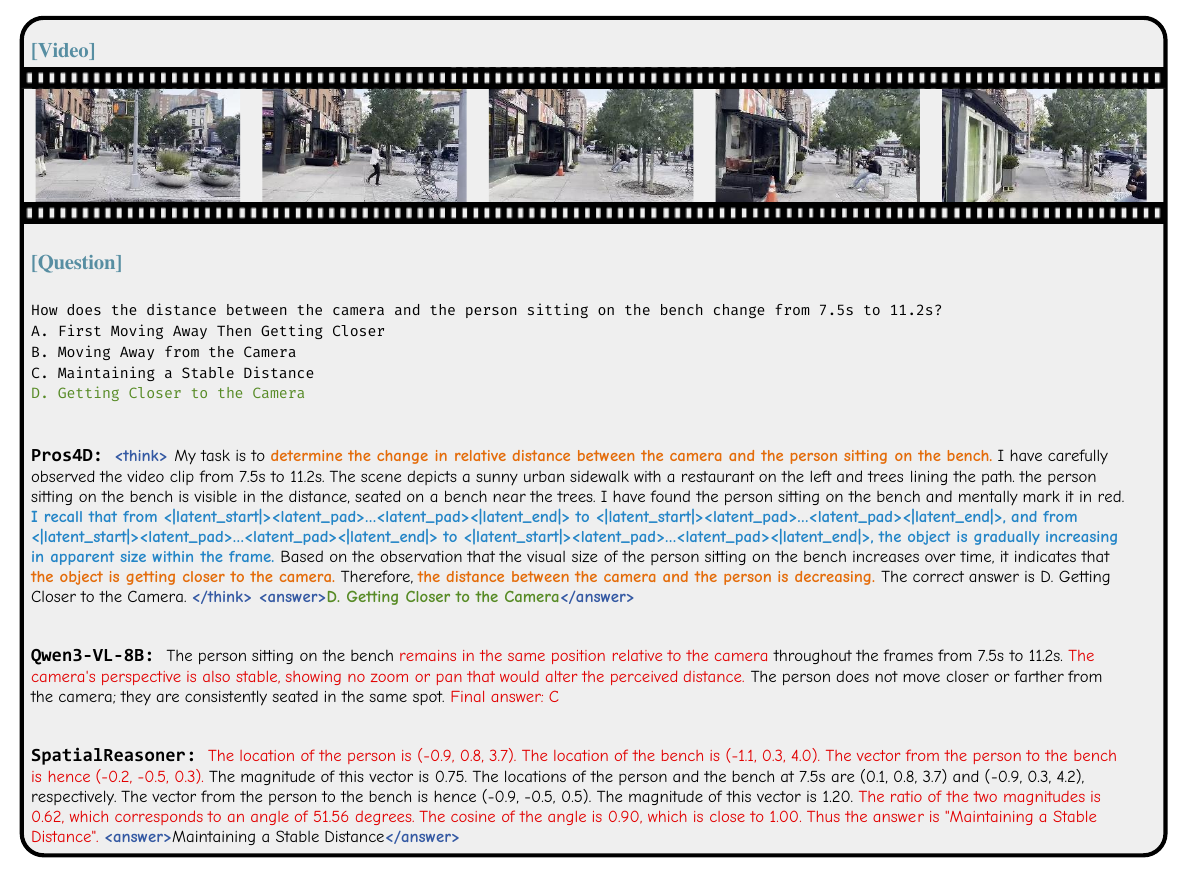}
\caption{\papertext{Camera-Object Distance on Pros4D-Bench. The comparison concerns the decreasing distance between the camera and the seated person.}{Pros4D-Bench 相机物体距离示例，比较对相机与坐着的人之间距离减小的理解。}}
\label{fig:app_pros_distance}
\end{figure*}

\begin{figure*}[p]
\centering
\includegraphics[width=\linewidth,height=0.88\textheight,keepaspectratio]{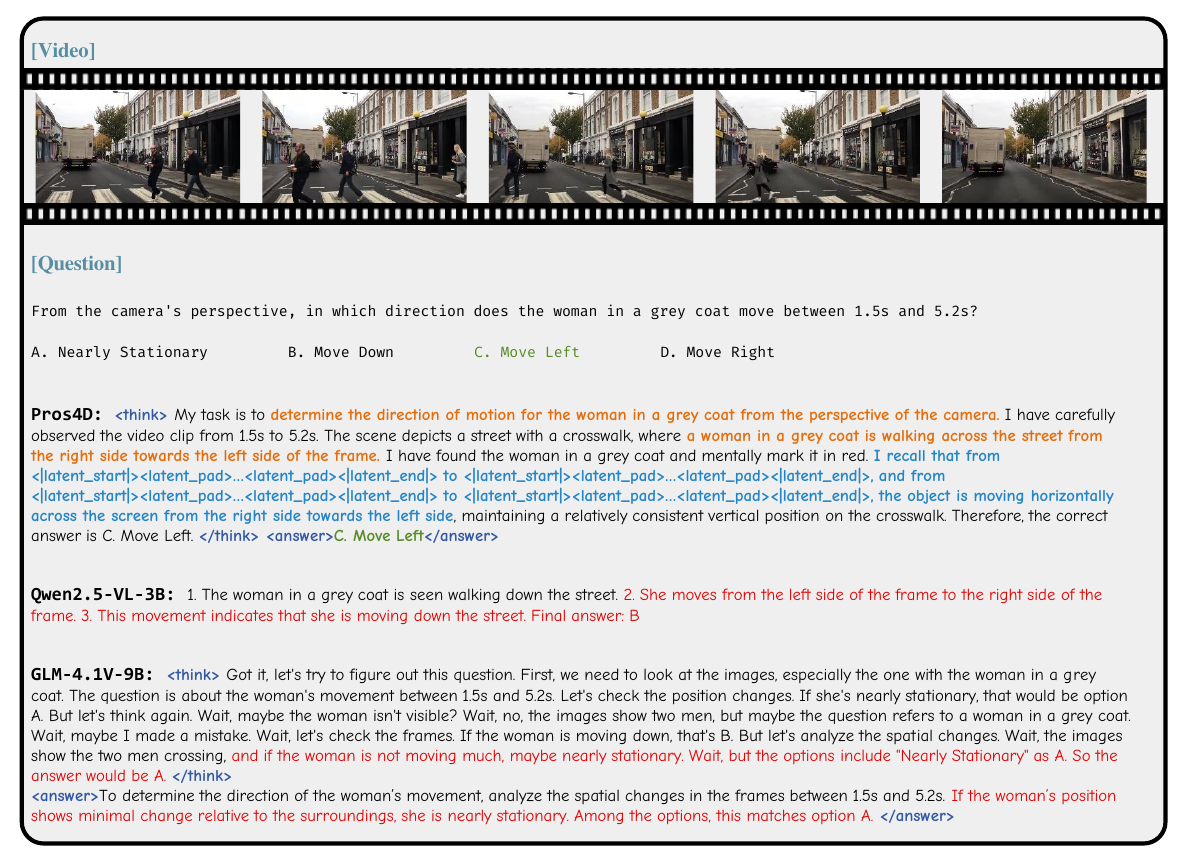}
\caption{\papertext{Object Motion Direction on Pros4D-Bench. The queried person moves leftward in the camera view.}{Pros4D-Bench 物体运动方向示例，被问行人在相机视角中向左运动。}}
\label{fig:app_pros_direction}
\end{figure*}

\begin{figure*}[p]
\centering
\includegraphics[width=\linewidth,height=0.88\textheight,keepaspectratio]{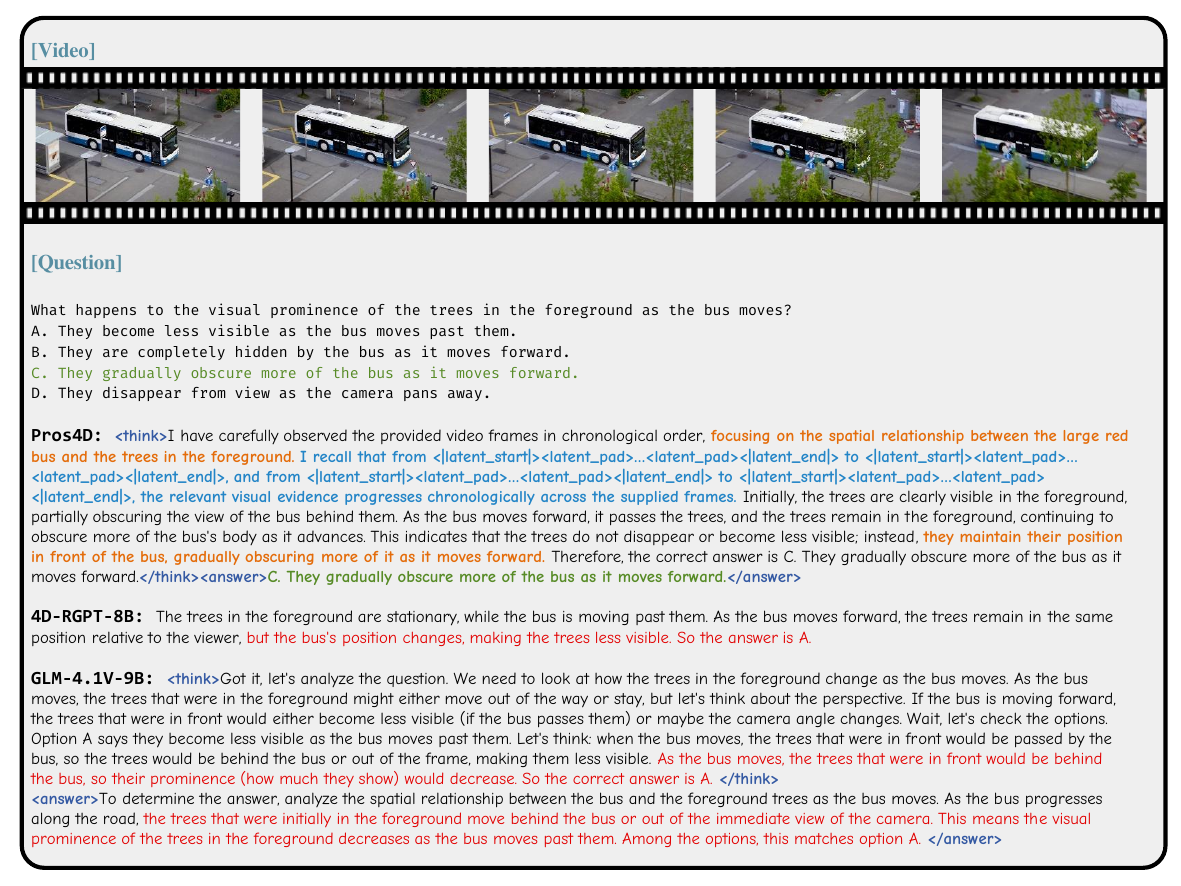}
\caption{\papertext{Object-scene relations on Dyn-Bench. The bus becomes increasingly occluded by trees as it moves through the scene.}{Dyn-Bench 物体场景关系示例，公交车运动过程中逐渐被树木遮挡。}}
\label{fig:app_dyn}
\end{figure*}

\begin{figure*}[p]
\centering
\includegraphics[width=\linewidth,height=0.88\textheight,keepaspectratio]{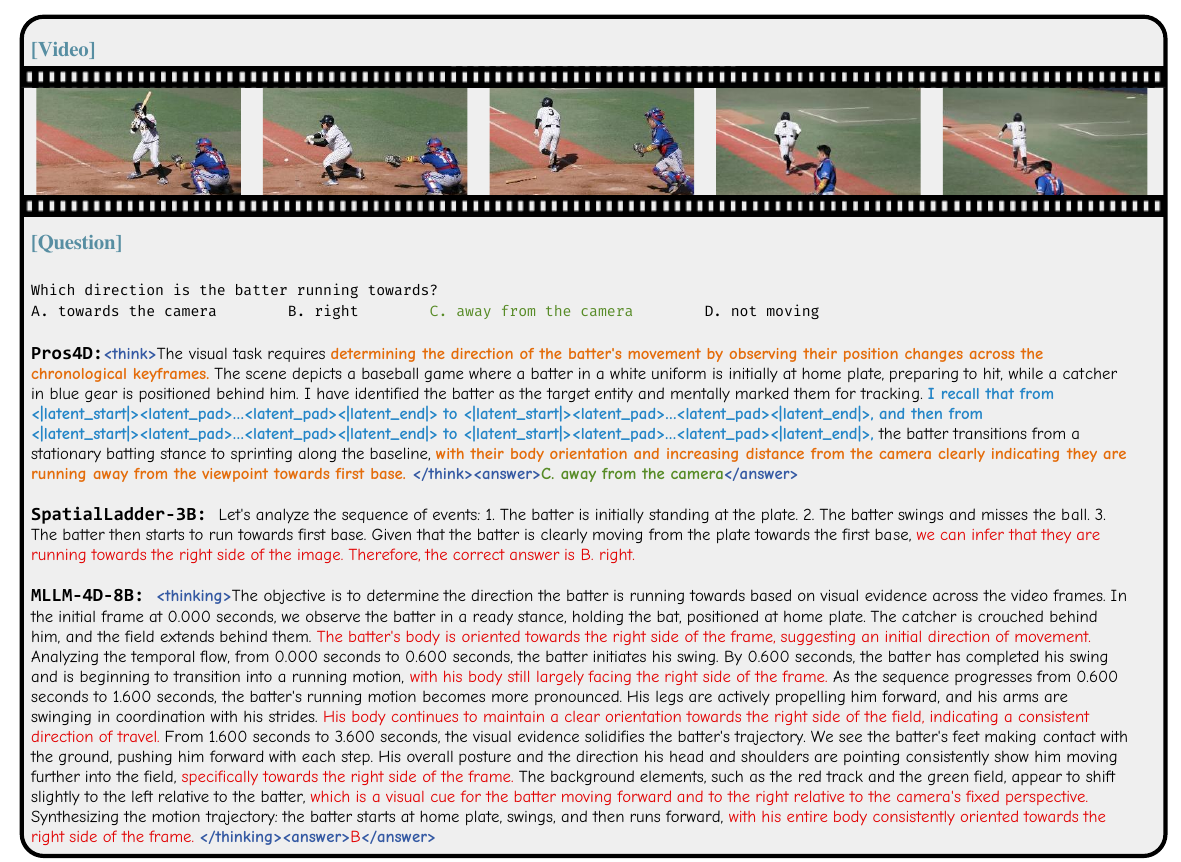}
\caption{\papertext{Observer-relative motion on VLM4D. The queried player moves away from the camera.}{VLM4D 相对观察者运动示例，被问运动员远离相机。}}
\label{fig:app_vlm4d}
\end{figure*}

\papertext{The qualitative comparisons illustrate how Pros4D relates motion to the queried entities. On Pros4D-Bench, our model recognizes the camera approaching a seated person where the baselines predict a stable distance (Figure~\ref{fig:app_pros_distance}), and identifies a pedestrian's leftward motion instead of downward or nearly stationary motion (Figure~\ref{fig:app_pros_direction}). Similar behavior appears on external videos, where Pros4D identifies increasing tree occlusion of a bus on Dyn-Bench (Figure~\ref{fig:app_dyn}) and a player moving away from the observer on VLM4D (Figure~\ref{fig:app_vlm4d}). These examples connect entity-specific states to observed-dynamics understanding of camera-relative motion, object movement, and scene relations. Continuous state vectors are omitted in the displayed traces; latent boundary and padding markers indicate their positions within the entity-specific reasoning trace.}{定性对比展示了 Pros4D 如何将运动与被问实体联系起来。在 Pros4D-Bench 上，模型识别出相机正接近坐着的人，而基线预测距离保持稳定（图~\ref{fig:app_pros_distance}）；模型还识别出行人向左运动，而非向下或近乎静止（图~\ref{fig:app_pros_direction}）。外部视频中也呈现相似行为，Pros4D 在 Dyn-Bench 上识别出树木对公交车的遮挡增加（图~\ref{fig:app_dyn}），并在 VLM4D 上判断球员正在远离观察者（图~\ref{fig:app_vlm4d}）。这些案例将实体相关状态与相机相对运动、物体移动及场景关系中的已观测动态理解联系起来。展示的轨迹省略了连续状态向量，潜在边界与填充标记表示它们在实体相关推理轨迹中的位置。}

\clearpage

\clearpage

\clearpage

\clearpage

\clearpage

\clearpage

\clearpage

\clearpage

\clearpage

\clearpage

\clearpage
\clearpage

\end{document}